%% file: main.tex
\input{macros}

\title{High-Accuracy List-Decodable Mean Estimation}
\author{Ziyun Chen\thanks{University of Washington, ziyuncc@cs.washington.edu. Supported by a Simons investigator award 928589, and an NSF grant CCF-2203541.} \and Spencer Compton\thanks{Stanford University, comptons@stanford.edu. Supported by the NDSEG Fellowship Program, Tselil Schramm’s NSF CAREER Grant no. 2143246, and Gregory Valiant’s Simons Foundation Investigator Award and NSF award AF-2341890.} \and Daniel Kane \thanks{University of California, San Diego, dakane@ucsd.edu. Supported by NSF Medium Award CCF-2107547.} \and Jerry Li\thanks{University of Washington, jerryzli@cs.washington.edu.}}

\begin{document}

\maketitle

\begin{abstract}
In list-decodable learning,  we are given a set of data points such that an $\alpha$-fraction of these points come from a ``nice'' distribution $D$, for some small $\alpha \ll 1$, and the goal is to output a short list of candidate solutions, such that at least one element of this list recovers some non-trivial information about $D$.
By now, there is a large body of work on this topic; however, while many algorithms can achieve optimal list size in terms of $\alpha$, all known algorithms must incur error which decays, in some cases quite poorly, with $1 / \alpha$.
In this paper, we ask if this is inherent: is it possible to trade off list size with accuracy in list-decodable learning?
More formally, given $\eps > 0$, can we can output a slightly larger list in terms of $\alpha$ and $\eps$, but so that one element of this list has error at most $\eps$ with the ground truth?
We call this problem \emph{high-accuracy list-decodable learning}.

Our main result is that non-trivial  high-accuracy guarantees, both information-theoretically and algorithmically, are possible for the canonical setting of list-decodable mean estimation of identity-covariance Gaussians.
Specifically, we demonstrate that there exists a list of candidate means of size at most $L = \exp \left( O\left( \tfrac{\log^2 1 / \alpha}{\eps^2} \right)\right)$ so that one of the elements of this list has $\ell_2$ distance at most $\eps$ to the true mean.
We also design an algorithm that outputs such a list with runtime and sample complexity $n = d^{O(\log L)} + \exp \exp (\widetilde{O}(\log L))$.
In particular, our results demonstrate that in the natural regime where $\alpha$ and $\eps$ are both small constants, it is possible to achieve error $\leq 0.01$ in fully-polynomial time, where all prior work suffered error which was much larger than $1$.
We do so by demonstrating a completely novel proof of identifiability, as well as a new algorithmic way of leveraging this proof without the sum-of-squares hierarchy, which may be of independent technical interest.
\end{abstract}

\section{Introduction}

Learning in the presence of outliers is a central challenge in statistics and machine learning. 
One of the most challenging, but important, formulations of this problem is the setting of \emph{list-decodable learning}~\cite{balcan2008discriminative,charikar2017learning}.
Here, the learner is given a dataset of $n$ points where all but an $\alpha$-fraction of the data is corrupted, for $\alpha \ll 1$, so that the vast majority of the dataset is complete noise.
In this setting, it is easy to see that it is generally impossible to recover the ground truth answer.
However, the influential work of~\cite{charikar2017learning} demonstrated that it is possible to output a short list of candidate solutions, one of which must achieve non-trivial closeness to the ground truth.

While these results are very interesting, they run into a fundamental bottleneck: namely, that the quantitative guarantees these estimates can attain are quite bad.
Consider the canonical setting of list-decodable mean estimation: given a dataset of points where an $\alpha$-fraction of the samples are drawn from a well-behaved distribution (say, an isotropic Gaussian) and the remaining $(1-\alpha)$-fraction are arbitrary outliers, all known efficient algorithms incur estimation error at least on the order of $\sqrt{\log(1/\alpha)}$. 
In particular, the statistical distance of the best estimate in the list and the ground truth Gaussian approaches $1$ as $\alpha$ goes to zero.
Moreover, it is known that such error is unavoidable, assuming the algorithm outputs a list of size at most $\poly (1 / \alpha)$~\cite{diakonikolas2023algorithmic}.

However, these lower bounds leave open an intriguing possibility: namely, that we could potentially achieve significantly higher accuracy---indeed, even arbitrarily good error---if we are willing to tolerate a slightly larger list size.
For instance, suppose that $\alpha$ is some small constant, say $\alpha = 0.25$.
Then, current algorithmic results would only seek to output a constant-sized list, and the error of the best estimator in this list would have error $\Omega (1)$.
But here, it is very reasonable to ask if, for $\eps > 0$ (say another small constant, e.g. $\eps = 0.01$), it is possible to output a list of slightly larger size that depends mildly on $1 / \eps$, so that the error of the best guess in the list is $\eps$.
We call this problem \emph{high-accuracy list-decodable mean estimation}.
Despite the wealth of work on list-decodable learning, there are no non-trivial guarantees, even information-theoretically, for this natural question.
Even basic questions in this vein are unresolved: for instance, prior to this work, it was not even known whether or not there exists a list of dimension-independent size which achieves such a guarantee.
In particular, all known identifiability proofs (both efficient and inefficient) for list-decodable learning fundamentally cannot achieve high-accuracy recovery guarantees.
Motivated by this discussion, we ask:
\begin{center}
    {\it What are the statistical and computational limits of high-accuracy list-decodable learning?}
\end{center}
Not only is this question natural in its own right, it also has immediate applications to the question of \emph{semi-verified learning}, first proposed by~\cite{charikar2017learning}.
Here, we are given a large dataset of noisy data, such that an $(1 - \alpha)$-fraction of it is arbitrary noise, and in addition, we are given a much smaller subset of $k$ trusted data points, which are guaranteed to be from the true distribution, and the goal is to synthesize the combined information to obtain better guarantees than are achievable with just the noisy or the trusted data points alone.
It is well-known that any algorithm for list-decodable learning implies non-trivial guarantees for this semi-verified setting, as we can use the list-decodable learning to output a small list of hypotheses, and then perform hypothesis selection on this list using our trusted data points.
In particular, doing so allows us to obtain non-trivial guarantees for semi-verified learning when the number of trusted points is much smaller than the dimension.
However, because previous algorithms for list-decodable learning could only obtain a low-accuracy list, the resulting semi-verified learner also suffered bad error as a result.
In contrast, if one could obtain a high-accuracy list-decodable learner, where the size of the list is sufficiently small, then one could hope to obtain significantly better error rates.
Indeed, the difficulty of optimally combining information with different levels of error (even in the special case of semi-verified learning) was observed in~\cite{chaudhuri2025robust}, where obtaining optimal statistical rates for semi-verified learning was posed as an interesting open question.

From a technical perspective, this question is also very interesting.
To somewhat oversimplify the state of affairs, all previous techniques for list-decodable learning  sought to recover a subset of points that shares $\alpha^2 n$ points with the true set of good points, usually with some additional regularity conditions on the recovered subset.
They then argued that this overlap, plus the regularity conditions, ensured that the statistics of the recovered set of points cannot deviate too far from the ground truth statistics.
However, it is not hard to see that, except in very special cases, such an argument fundamentally cannot obtain high-accuracy guarantees, because it cannot distinguish between the ground truth distribution, and the ground truth distribution conditioned on an event of probability $\alpha$, and the statistics of these two events can differ wildly.
This is to all to say that conceptually novel ideas are necessary to obtain any high-accuracy guarantees in the list-decodable learning setting.

\subsection{Our Results}
In this work, we obtain the first non-trivial guarantees for high-accuracy list-decodable learning.
Specifically, we consider the canonical setting of list-decodable mean estimation for isotropic Gaussians in $d$-dimensions.
Formally, we consider the following, standard noise model for list-decodable estimation:
\begin{definition}
    We say a set of points $S$ is an \emph{$\alpha$-pure set} of points with respect to a distribution $D$ if it contains a subset $\Sgood \subset S$ so that $|\Sgood| = \alpha |S|$, and $\Sgood$ is a collection of independent samples from $D$.
\end{definition}
Note that we make no assumptions about the other points in $S$; for instance, they could even be chosen adversarially depending on the points in $\Sgood$.
We also note that this is also equivalent to saying that $S$ is a $(1 - \alpha)$-additively corrupted dataset in the terminology of~\cite{diakonikolas2018list}, however, we find that for this setting it is slightly more intuitive to measure the fraction of inliers rather than the fraction of outliers.

\begin{definition}[List-decodable Gaussian mean estimation]
\label{def:main}
    Let $\alpha, \eps > 0$, and let $\mu \in \mathbb{R}^d$.
    Given a dataset $\alpha$-pure dataset $S$ with respect to $N(\mu, I)$ of size $n$, output a list of $L$ candidate means $\mu_1, \ldots, \mu_L$ so that with high probability,
    \[
    \min_{i} \norm{\mu_i - \mu}_2 \leq \eps \; .
    \]
    We refer to this quantity $\eps$ as the \emph{error} of the list-decodable learning algorithm.
\end{definition}
\noindent
Our first result is a tight characterization of the information-theoretic limits of this problem.
Specifically, we show:
\begin{theorem}[informal, see \cref{thm:info-theory,lemma:lb}]
\label{thm:inefficient-informal}
    In the setting of Definition~\ref{def:main}, there is an (inefficient) estimator which, for $n$ sufficiently large, outputs a list of size 
    \[
    L = \exp \left(O \left( \frac{\log^2 (1 / \alpha)}{\eps^2} \right) \right)
    \]
    candidate means, which achieves error $\eps$ with high probability.
    Moreover, any algorithm which achieves error $\eps$ with constant probability must output a list of size $\exp \left( \Omega \left( \tfrac{\log^2 (1 / \alpha)}{\eps^2} \right) \right)$.
\end{theorem}
\noindent
We pause to make several remarks about this result.
First, note that the list size is completely independent of the dimension.
Second, we observe that in the aforementioned setting where $\alpha, \eps$ are both small constants, this is the first result that demonstrates that error which is significantly smaller than a large constant larger than $1$ is possible for list-decodable mean estimation.
The key technical idea is a new identifiability proof which directly argues that the set of possible candidate means must have bounded size, using ideas from Gaussian process theory and isoperimetry.
See \cref{subsec:info-overview,sec:info} for a more in-depth discussion of these ideas.

The estimator which achieves the upper bound in Theorem~\ref{thm:inefficient-informal} is inefficient, and moreover, works in the asymptotic setting, i.e. when $n$ could be very large.
Our second result is a new, efficient estimator with non-asymptotic guarantees, which achieves the same error:
\begin{theorem}[informal, see \cref{thm:efficient}]
    In the setting of Definition~\ref{def:main}, there is an algorithm which outputs a list of size at most
    \[
    L = \exp \left(O \left( \frac{\log^2 (1 / \alpha)}{\eps^2} \right) \right)
    \]
    candidate means, which achieves error $\eps$ with probability at least $0.99$ when the sample size $n \ge d^{O(\log L)} + \exp(L)$. The time complexity is $d^{O(\log L)} + \exp\exp(\tilde{O}(\log L))$.
\end{theorem}
In particular, in the regime where $\alpha, \eps$ are small constants, our runtime and sample complexity are fully polynomial, demonstrating that it is possible to achieve small constant error efficiently in the list-learning setting.

By combining this with standard hypothesis selection routines, an immediate implication of this is a new algorithm that achieves high-accuracy guarantees for semi-verified learning, with very few trusted points:
\begin{corollary}\label{cor:semiverified}
Let $\alpha, \eps > 0$, and let $\mu \in \mathbb{R}^d$.
Suppose we are given an $\alpha$-pure dataset $S_{\mathrm{noisy}}$ for $N(\mu, I)$ of size $n_1$, as well as a dataset of $S_{\mathrm{trusted}}$ of $n_2$ points drawn independently from $N(\mu, I)$.
Suppose that 
\[
n_1 \geq d^{O(\log(L))} + \exp(L) \; , n_2 \geq \Omega \left( \frac{\log L}{\eps^2} \right) \; ,
\]
Then, there is an algorithm which runs in time $d^{O(\log L)} + \exp\exp(\tilde{O}(\log L))$, and which outputs $\widehat{\mu}$ so that $\norm{\widehat{\mu} - \mu}_2 \leq \eps$ with probability at least $0.99$.
\end{corollary}
We outline a short proof in \cref{app:semiverified}.

From a technical perspective, an interesting aspect of our algorithm is that it once again departs from the ``standard'' paradigm for algorithm design in the list-learning setting.
To once again oversimplify, typically, after one has established a complex identifiability proof such as Theorem~\ref{thm:inefficient-informal}, the ``standard'' approach is to use techniques from SDP hierarchies such as the sum-of-squares hierarchy to convert the proof into an efficient algorithm, see e.g.~\cite{kothari2018robust,karmalkar2019list,raghavendra2020list,diakonikolas2022list,bakshi2021list,ivkov2022list}.
However, our identifiability proof uses somewhat sophisticated machinery, including the previously mentioned Gaussian isoperimetry, and does not seem to easily lift into SoS.

Instead, we propose a two-step algorithm.
First, we efficiently identify a low-dimensional subspace which must essentially contain all possible candidate means, by using a new filtering algorithm based on high-degree Hermite polynomials, which may be of independent technical interest.
Then, within this subspace, we are able to prune the list of possible candidate means down to the correct size.
Crucially, this second step appeals to our identifiability theorem (Theorem~\ref{thm:inefficient-informal}) in a black-box fashion, to demonstrate that the number of candidates our pruning procedure can return is small.
See \cref{subsec:efficient-overview,sec:efficient} for a more detailed description of our algorithm.

\subsection{Related work}
List-decodable learning was first proposed in work of~\cite{balcan2008discriminative,charikar2017learning}.
By now there is a rich literature on the topic, including efficient algorithms for list-decodable mean estimation~\cite{kothari2018robust,diakonikolas2018list,raghavendra2020list,diakonikolas2021list, diakonikolas2022clustering}, sparse mean estimation~\cite{zeng2022list, diakonikolas2022list}, covariance estimation~\cite{ivkov2022list}, linear regression~\cite{karmalkar2019list,raghavendra2020list,bakshi2021list,das2023efficient}, and even more general settings~\cite{charikar2023characterization,klivans2025power}, as well as computational lower bounds~\cite{diakonikolas2021statistical}.
However, none of these works achieve high-accuracy guarantees similar to the ones we are interested in.
List-decodable learning is also closely related to the larger literature on robust statistics~\cite{huber1992robust,tukey1960survey,anscombe1960rejection,tukey1975mathematics}, and in particular, the recent wave of interest in algorithmic robust statistics, beginning with work of~\cite{diakonikolas2019robust,lai2016agnostic}; we defer the interested reader to~\cite{diakonikolas2023algorithmic} for a more detailed description of this literature.

As described above, list-decodable learning is also closely related to the semi-verified learning problem~\cite{charikar2017learning, meister2018data,zeng2023semi,chaudhuri2025robust}, however, none of these results bear any technical relevance to our setting.

\section{Preliminaries}
Throughout this work, for conciseness of exposition, $c>0$ will denote any sufficiently small constant, and $C>0$ will denote any sufficiently large constant. 
In different lines, the value of $c,C$ may change.
For a matrix $M$, we will let $\norm{M}_2$ denote its spectral norm, and we will let $\norm{M}_F$ denote its Frobenius norm. Unless the base is otherwise specified, $\log$ refers to the natural log.

We will also require the following classic result:
\begin{theorem}[Sudakov's Minoration Inequality~\cite{sudakov1969gaussian}]\label{thm:sudakov-minoration}
    For a mean zero Gaussian process, for any $\eps \ge 0$ we have
    \begin{equation*}
        \Ex\left[\sup_{i\in [l]} X_t \right]\ge c \eps \sqrt{\log(\mathcal{N}(T,d,\eps))}
    \end{equation*}
\end{theorem}

\subsection{Low-accuracy list-decodable mean estimation}
As a preprocessing routine, we will first use the previous work that obtains efficiently obtains low-accuracy list learning guarantees but with a small list size.
Specifically, we will need:
\begin{theorem}[\cite{diakonikolas2022list}]
\label{thm:low-accuracy}
    Let $\alpha > 0$.
    There is an algorithm which, given an $\alpha$-pure dataset $S$ with respect to $N(\mu, I)$ of size $n = d^{O(\log (1 / \alpha))}$, outputs a list of $L = O(1 / \alpha)$ candidate means $\mu_i, \ldots, \mu_L$ so that with probability at least $0.99$, there exists $i \in [L]$ so that $\norm{\mu_i - \mu}_2 \leq O(\sqrt{\log 1 / \alpha})$.
\end{theorem}

\section{Technical overview}
The main contributions of our paper are an information-theoretic proof of the near-optimal list size, and an accompanying efficient algorithm. In this section, we highlight the key ideas for both.

\subsection{Information-theoretic bound}\label{subsec:info-overview}
In this section, we will discuss the existence of a small list for high-accuracy list-decodable learning in an infinite-sample setting. These techniques will later yield desired guarantees for the typical finite-sample list-decodable setting.

Consider the infinite-sample regime for list-decodable learning: the inliers will be distributed according to the distribution $N(\mu,I)$, and the only power of the adversary is where to add the rest of the mass. Hence, in the infinite-sample limit, we expect to see a distribution $D$ where $D(x) \ge \alpha [N(\mu,I)](x) \, \, \forall x \in \mathbb{R}^d$. When we observe such a distribution, there may be many values of $\mu \in \mathbb{R}^d$ for which a distribution $D$ satisfies this condition. We similarly define a notion for whether a potential mean $\mu \in \mathbb{R}^d$ is consistent with plausibly being an $\alpha$-fraction of $D$:

\begin{definition}[$\alpha$-consistent]\label{def:consistent}
    $\mu \in \mathbb{R}^d$ is $\alpha$-consistent if $D(x) \ge \alpha \cdot [N(\mu,I)](x) \, \, \forall x \in \mathbb{R}^d$.
\end{definition}

We desire to output a small list where every $\alpha$-consistent $\mu$ is close to an item in our list. Our procedure for inefficiently constructing a list for $D$ is simple: while there exists an $\alpha$-consistent $\mu$ that is $\eps$-separated from all previous items in our list, then add any such $\mu$. In this sense, our task boils down to the core question: what is the maximum-size list $\mu_1,\dots,\mu_l$ of $\alpha$-consistent and $\eps$-separated $\mu_i$?

\textbf{Warm-up: sketch for $\eps = 10 \sqrt{\log(1/\alpha)}$.} When $\eps$ is large enough, the work of Diakonikolas, Kane, and Stewart \cite{diakonikolas2018list} yields a simple yet illuminating answer to our core question. Here we present a proof implied by theirs. For sake of contradiction, consider a list $\mu_1,\dots,\mu_l$ of $\alpha$-consistent and $10 \sqrt{\log(1/\alpha)}$-separated $\mu_i$, where $l = \frac{4}{\alpha}$. Let us define the regions $R_1,\dots,R_{l}$ that correspond to regions where the density of $N(\mu_i,I)$ is larger than all other $N(\mu_j,I)$. Without proof, we give intuition that because the $\mu_i$ are all well-separated, at least half of the mass of each $N(\mu_i,I)$ must be in the region $R_i$ where its density is the largest:
\begin{equation*}
    \Pr_{X \sim N(\mu_i,I)}[X \in R_i] \ge  1 - \sum_{j \ne i} \Pr_{X \sim N(\mu_i,I)}\left[ [N(\mu_i,I)](x) \le [N(\mu_i,I)](x) \right] \ge \frac{1}{2}
\end{equation*}
Yet, by definition of $\alpha$-consistency, the density of $D$ must always be at least $\alpha$-fraction the density of any $N(\mu_i,I)$. We are then able to conclude a contradiction that the total mass of $D$ must exceed $1$, since too much mass must be in each region $R_i$:
\begin{equation*}
    1 \ge \sum_{i=1}^l \Pr_{X \sim D}[X \in R_i] \ge \sum_{i=1}^l \alpha \Pr_{X \sim N(\mu_i,I)}[X \in R_i] \ge \frac{\alpha \cdot l}{2} = 2
\end{equation*}
Accordingly, this proves that any list $\mu_1,\dots,\mu_l$ of $\alpha$-consistent and $10 \sqrt{\log(1/\alpha)}$-separated $\mu_i$ must have size at most $O(1/\alpha)$.

In the actual proof of \cite{diakonikolas2018list}, the analysis mostly goes as we have just described. Their conceptual emphasis is less explicitly about geometric regions $R_i$, and more about how each $\mu_i$ has an associated set, and these sets have limited overlap. In our upcoming proof, we will crucially use how our $R_i$ are the \textit{Voronoi cells}, meaning the regions where $\mu_i$ has the largest density. 

\textbf{Intuition for our result. } Recall how $R_i$ correspond to the regions where $N(\mu_i,I)$ has the largest density, meaning $R_i = \{ \left\|x-\mu_i\right\|^2 < \left\|x-\mu_j\right\|^2, \forall j \neq i \}$. Let $q_i = \Pr_{x\sim N(\mu_i,I)}[ X \in R_i]$. In this language, our analysis will follow a similar program to the warm-up but with sharper guarantees: we will closely analyze the $q_i$, and then similarly conclude the sum of $q_i$ must exceed $1/\alpha$ if there are too many $\alpha$-consistent $\mu_i$ that are $\eps$-separated.

Our first observation is that we find it helpful to localize our analysis to small balls. We use the initial warm-up result to reduce our task to $O(1/\alpha)$ subproblems, where after re-centering we may assume $\mu$ is in a bounded ball such that $\| \mu \|_2 \le O(\sqrt{\log(1/\alpha)})$. 

We now introduce the following auxiliary probability sequence $(p_i)_{i\in [l]}$, defined as $p_i = \Pr_{x\sim N(0,I)}[x\in A_i]$, where $A_i = \{x: \langle x, \mu_i\rangle > \langle x,\mu_j \rangle, \forall j\neq i\}$. This region $A_i$ will be a surprisingly insightful reference for analyzing the mass in regions $R_i$.

Let $R'_i$ correspond to the region $R_i$ translated by $-\mu_i$. By definition,

\begin{equation*}
    q_i = \Pr_{X \sim N(\mu_i,I)}[X \in R_i] = \Pr_{X \sim N(0,I)}[X \in R_i']
\end{equation*}

\begin{figure}[t]
  \centering
  \begin{minipage}{0.48\textwidth}
    \centering
    \includegraphics[width=\linewidth]{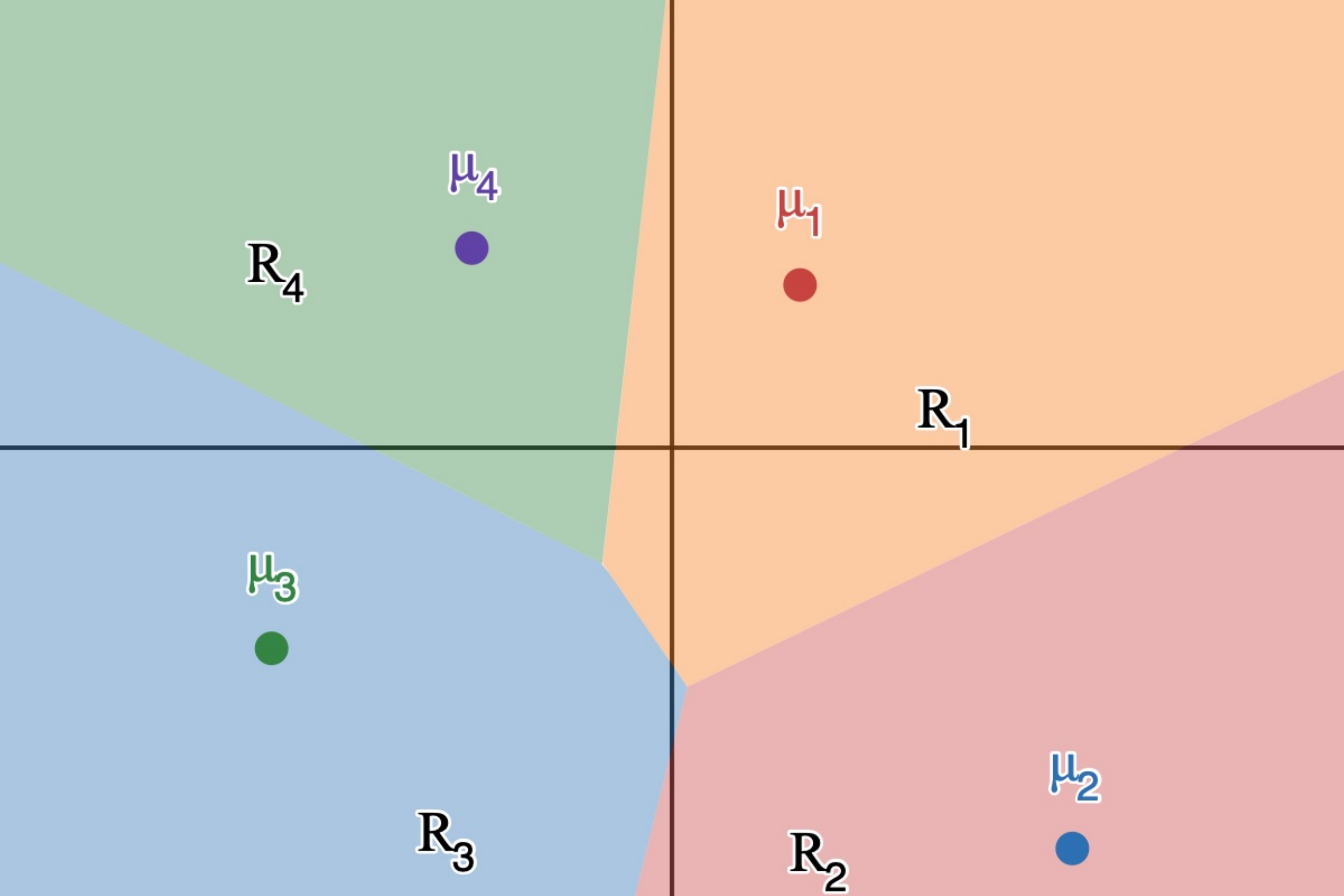}
  \end{minipage}\hfill
  \begin{minipage}{0.48\textwidth}
    \centering
    \includegraphics[width=\linewidth]{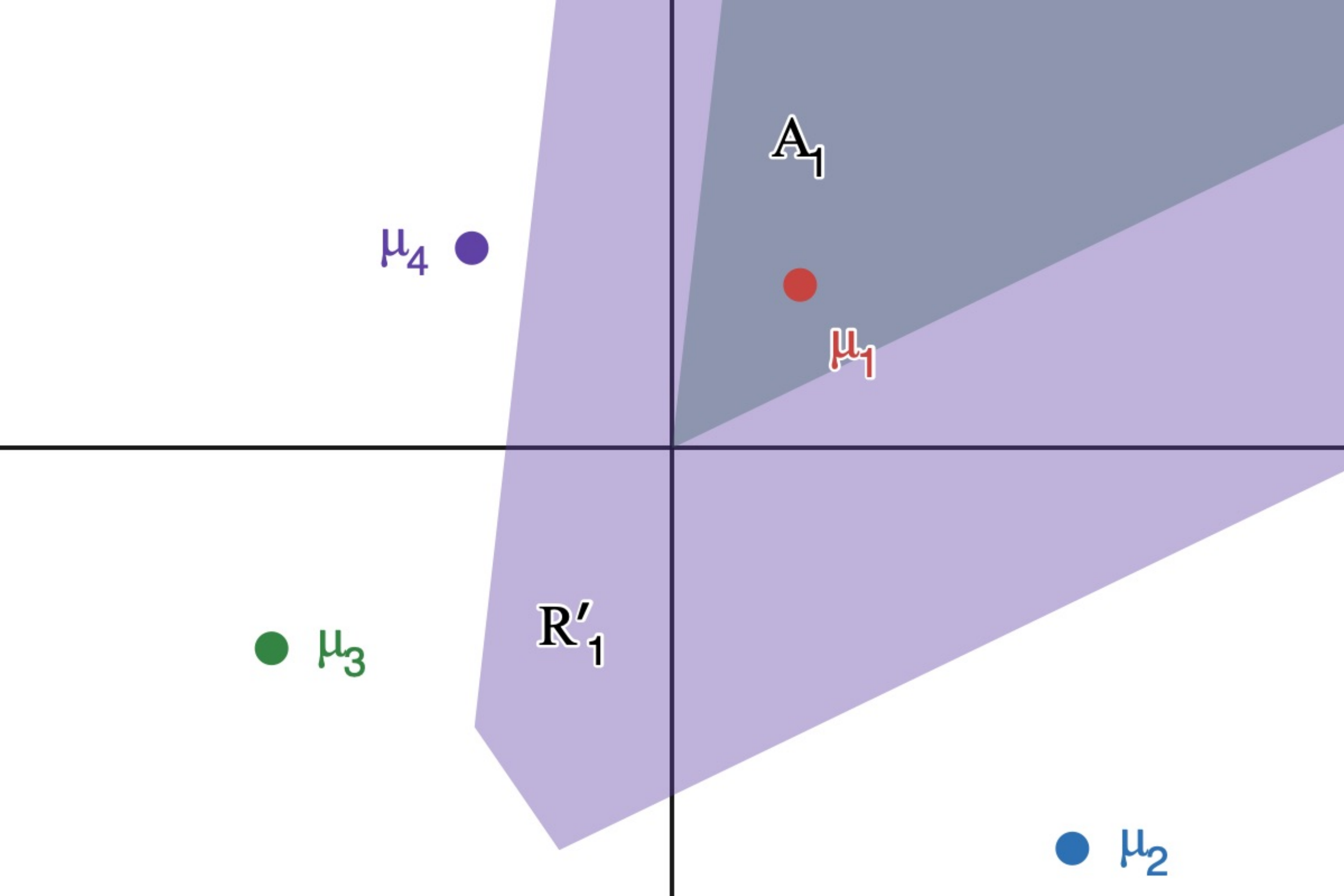}
  \end{minipage}
  \caption{The left figure illustrates the regions $R_i$ where their density is the maximum for the points $\mu_1,\mu_2,\mu_3,\mu_4$. The right figure focuses on $A_1$ and $R_1'$, illustrating how $R_1'$ contains a fattening of $A_1$. Since $q_i \ge \Pr_{X \sim N(0,I)}[X \in R_1']$, this perspective will enable lower bounds for $q_i$ in terms of $A_i$.}
  \label{fig:regions}%
\end{figure}

This indicates how relationships between $A_i$ and $R_i'$ may inform the value of $q_i$. We initially observe that $q_i \ge p_i$ via the following simple analysis that $R_i' \supseteq A_i$:
\begin{equation*}
    R_i' = \{x : \left\|x\right\|^2 < \left\|x+(\mu_i-\mu_j)\right\|^2, \forall j \neq i\} \supseteq \{ x : \langle x, \mu_i \rangle > \langle x, \mu_j \rangle, \forall j \neq i\} = A_i
\end{equation*}

More consequentially, after a short calculation, we observe $R_i'$ actually contains an $\eps/2$-\textit{fattening} of $A_i$ (e.g. see \cref{fig:regions}). Combined with the Gaussian isoperimetry inequality, this will give us a favorable lower bound for $q_i$ in terms of $p_i$. For example, in the regime where $\eps \in (0,1]$, we use
\begin{equation}
    q_i \ge  p_i \cdot \exp(c\epsilon \sqrt{\log(1/p_i)}).\label{eq:iso-example}
\end{equation}

Let us describe intuition for how this bound leads us towards a proof. For simplicity, consider a list $\mu_1,\dots,\mu_l$ where the auxiliary probabilities $p_i$ happen to be uniform: $p_1,\dots,p_l = \frac{1}{l}$. This is not generally true, but is an insightful example. In this case, using \cref{eq:iso-example}, we could conclude that for any list of size $l > \exp(C \log^2(1/\alpha)/\eps^2)$, the sum of $q_i$ must exceed $1/\alpha$ and cause a contradiction. 

Of course, the auxiliary probabilities $p_i$ need not be uniform. Moreover, the lower bound of \cref{eq:iso-example} is not fruitful in many cases, such as an example where $(p_1,\dots,p_l) = (\frac{1}{2},\frac{1}{4},\frac{1}{8},\dots)$. A key observation, however, is that the values of $p$ cannot be arbitrary. Indeed, $p_1,\dots,p_l$ correspond to the distribution of the $\argmax$ for a Gaussian process $\sup_{i \in [l]} \langle x, \mu_i \rangle$. With a careful technical argument, we show that the distribution $p$ must be well-spread enough for \cref{eq:iso-example} to yield a favorable bound, or otherwise the distribution $p$ would cause a violation of Sudakov's minoration inequality (and hence is not consistent with the distribution of the optimizer of a Gaussian process). Eventually, this yields our desired information-theoretic guarantee.

\subsection{Efficient list-decodable learning}\label{subsec:efficient-overview}
Our plan will be to design an algorithm that efficiently leverages our information-theoretic proof. In that proof, our procedure was to simply choose $\eps$-separated, $\alpha$-consistent $\mu_i$ repeatedly until there were no more such $\mu_i$. For this section, it is both impossible to know whether a $\mu_i$ is $\alpha$-consistent from only finite samples, and also non-obvious how to efficiently find $\mu_i$ with our desired properties. 

As a remedy for this, we will choose $\mu_i$ for which there exists a collection of $\alpha n$ samples, where these samples centered around $\mu_i$ have empirical moments that are very close to the standard Gaussian $N(0,I)$. Relaxing this slightly, we may look for a $\mu_i \in \mathbb{R}^d$ and sample weights $w_1,\dots,w_n \in [0,1]$ with $\sum_i w_i = \alpha n$, where $\mu_i$ is $\eps$-separated from the previously chosen list entries, and 
\begin{equation*}
    \frac{1}{\alpha n} \sum_i w_i \cdot \langle X_i - \mu_i, v \rangle^j \approx \Ex_{X \sim N(0,1)}[x^j]
\end{equation*}
for all $\|v\|_2 = 1$ and $j \in \{1,\dots,k\}$ for some bounded $k$. 

First, we will employ prior work of \cite{diakonikolas2022list} to efficiently reduce our task to $O(1/\alpha)$ subproblems, where each has the restriction $\| \mu_i \|_2 \le O(\sqrt{\log 1/\alpha})$. 

\textbf{Finding a low-dimensional subspace.} Next, we will find a low-dimensional subspace that is close to the true $\mu$ with high probability. Crucially, the dimension of this subspace will not depend on $d$. 
We will do so via a two-step process.
The first step will be an iterative filtering algorithm similar to those in the robust statistics literature~\cite{diakonikolas2019robust} on the empirical high-order Hermite polynomial tensors of the dataset.
More specifically, let $H_t (x): \mathbb{R}^d \to \mathbb{R}^{d^t}$ denote the $t$-th order (probabilist's) Hermite polynomial tensor (see Definition~\ref{def:hermite}), for $t$ appropriately chosen but independent of $d$.
By standard results in high-dimensional probability, we know that if we take enough samples, the empirical statistics over the good samples of the Hermite polynomials concentrate very tightly around the population statistics, and the population statistics of the Hermite polynomials satisfies
\[
\norm{\Ex_{X \sim N(\mu, I)} \left[ H_t (X) \otimes H_t (X) \right] }_2 \leq t! \cdot C^t \exp \left( \sqrt{t} \norm{\mu}_2 \right) \; ,
\]
which is in particular, much smaller than $\exp (L)$ for the list size $L$ we are targeting.
We show that these facts imply that we can use iterative filtering to remove samples from the dataset so that (1) we remove almost no good points, and (2) the $\ell_2$-norm of the empirical Hermite polynomial (when treated as a $d^t$-length vector) over the remaining data points is bounded.
See Lemma~\ref{lem:filter-main} for the full analysis.
Once we have this, the $\ell_2$ bound on the degree $t$ empirical Hermite polynomial implies that if we flatten it into a $(d \times d^{t - 1})$ matrix, this matrix must have bounded Frobenius norm, and so in particular, it has a small number of large singular values.
As a final step, we demonstrate that if we take the union of the span of the left singular vectors with large singular values of these flattened matrices, this subspace must approximately contain $\mu$.
Intuitively, this is because our condition implies that in all directions orthogonal to this subspace, the the low-degree moments of dataset match that of a standard normal Gaussian.
However, if there was a candidate mean that was not contained in this subspace, then this would necessarily induce a large moment, and thus we can conclude that no such candidate mean can exist.

\textbf{Searching for moment-matching candidates.} With a low-dimensional subspace in hand, we may now afford running times with exponential dependence in the dimension of the subspace. Thus, we may exhaustively search over a net for points that approximately match moments as desired. Checking whether a fixed center has sample weights $w_1,\dots,w_n$ that match moments is an efficiently-solvable convex program. This subroutine produces an $\eps/2$-separated list of points $\mu_1,\dots,\mu_l$ satisfying our moment-matching condition, and at least one of these points will be $\eps$-close to the true $\mu$ with high probability.

\textbf{Bounding list size via fooling.} The main remaining question is whether this list $\mu_1,\dots,\mu_l$ must be small. We will show any collection of $\eps/2$-separated $\mu_i$ that satisfy this moment condition must have bounded size. Recall $R_i \triangleq \{ \|x - \mu_i \| < \|x - \mu_j \| \}$: the region where $N(\mu_i,I)$ has the largest density. In our information-theoretic proof, we showed that in the case where the $\eps$-separated $\mu_i$ were all $\alpha$-consistent, then for large enough $l$ there is a contradiction
\begin{equation*}
    1 \ge \sum_{i=1}^l \Pr_{X \sim D}[  X \in R_i] \ge \sum_{i=1}^l \alpha \cdot \Pr_{X \sim N(\mu_i,I)}[X \in R_i] > 1.
\end{equation*}
In the original proof, most of the difficulty was in the last step of the inequality. When we adapt this argument for moment-matching $\mu_i$, the main difficulty is the second inequality. For any moment-matching $\mu_i$, let $w^{(i)}$ be the corresponding vector of sample weights in $[0,1]$, and let $D_i$ be the normalized empirical distribution over these samples: $D_i = \frac{1}{\alpha n}\sum_{j=1}^n w_{j}^{(i)} \cdot X_j$. Our same information-theoretic proof will show that our list of moment-matching $\mu_i$ cannot be too large, if we prove
\begin{equation*}
    \sum_{i=1}^l \alpha \cdot \Pr_{X \sim D_i}[X \in R_i]  \approx \sum_{i=1}^l \alpha \cdot \Pr_{X \sim N(\mu_i,I)}[X \in R_i].
\end{equation*}
Meaning, we must prove the normalized empirical distribution from $w^{(i)}$ has approximately as many points in $R_i$ as the distribution $N(\mu_i,I)$. In general, just because the lower-order moments of $D_i$ match the moments of $N(\mu_i,I)$, does not mean the proportion of their samples within some region will be approximately the same. However, observe that our particular region $R_i$ is exactly the intersection of of $l-1$ halfspaces. There is a rich body of work in pseudorandomness (e.g. \cite{bazzi2009polylogarithmic,braverman2008polylogarithmic,diakonikolas2010bounded2,diakonikolas2010bounded,klivans2013moment,kane2013learning}) and more recently, testable learning (e.g. \cite{rubinfeld2023testing,gollakota2023moment,diakonikolas2023efficient,klivans2024testable,slot2024testably,diakonikolas2024testable}) studying how matching moments implies fooling concept classes like halfspaces; hence, distributions that match moments will have roughly the same proportion of samples within an intersection of halfspaces. The works of \cite{diakonikolas2010bounded,gollakota2023moment} guide us towards the technical fooling result we desire. However, we cannot immediately use either of these results without modification, since our application only matches moments \textit{approximately} (the statement of \cite{diakonikolas2010bounded} uses exact matching), and we desire tighter guarantees for super-constant numbers of halfspaces than given by \cite{gollakota2023moment}. 

All together, since we efficiently find a list $\mu_1,\dots,\mu_l$ where each $\mu_i$ has a corresponding sample weights $w^{(i)}$ that match moments with $N(\mu_i,I)$, and since moment-matching fools the intersection of halfspaces, then our information-theoretic proof bounds the size of our list.

\section{Information-theoretic bound}\label{sec:info}
In this section, we will show the existence of a small list for high-accuracy list-decodable learning. We will discuss an infinite-sample setting, but this will later give us results that yield the desired guarantee for the typical list-decodable setting.

Recall the notion of $\alpha$-consistency (\cref{def:consistent}) describing whether a potential mean $\mu \in \mathbb{R}^d$ is consistent with plausibly being an $\alpha$-fraction of $D$. We now state our main information-theoretic result for list-decodable mean estimation:

\begin{theorem}\label{thm:info-theory}
    Consider any distribution $D$ over $\mathbb{R}^d$, inlier parameter $\alpha \in (0,\nicefrac{1}{2}]$, and accuracy parameter $\eps \in (0,10 \sqrt{\log(1/\alpha)}]$. There exists a list $L$ such that for any $\alpha$-consistent $\mu \in \mathbb{R}^d$, there is a $\mu' \in L$ such that $\|\mu - \mu'\|_2 \le \varepsilon$. Further, $L$ has size at most $\exp(C \log^2(1/\alpha)/\eps^2)$ for sufficiently large $C>0$.
\end{theorem}

This is an information-theoretic version of our eventual goal; we will not worry about leveraging only a finite number of samples or having desirable runtime.

Diakonikolas, Kane, and Stewart \cite{diakonikolas2018list} showed that you may learn a list of size $O(1/\alpha)$ with accuracy $\|\mu - \mu'\|_2 \le O(\sqrt{\log(1/\alpha)})$. Our improvement in this work is improving the accuracy beyond this barrier. The information-theoretic version of the classical result is quite simple, and we will present a sketch of a proof implied by theirs as a warm-up.

\subsection{Warm-up: sketch for learning up to accuracy $O(\sqrt{\log(1/\alpha)})$ }
\begin{lemma}\label{lemma:classic}
    Consider any distribution $D$ over $\mathbb{R}^d$ and an inlier parameter $\alpha \in (0,\nicefrac{1}{2}]$. There exists a list $L$ such that for any $\alpha$-consistent $\mu \in \mathbb{R}^d$, there is a $\mu' \in L$ such that $\|\mu - \mu'\|_2 \le 10 \sqrt{\log(1/\alpha)}$. Further, $L$ has size at most $\frac{4}{\alpha}$.
\end{lemma}
\begin{proof}
Suppose there existed a list of means $L = [\mu_1,\dots,\mu_{l}]$, where all $\mu_i$ are $\alpha$-consistent, and all pairs are $10 \sqrt{\log(1/\alpha)}$-separated (meaning, $\| \mu_i - \mu_j \|_2 \ge 10 \sqrt{\log(1/\alpha)}$ for $i \ne j$). We will show by contradiction that it is impossible for such a list to have size $\lceil 4/\alpha\rceil $. 

Let us define regions $R_1,\dots,R_{l}$ that correspond to regions where the density of $N(\mu_i,I)$ is larger than all other $N(\mu_j,I)$ for $i \ne j$. 

First, we observe how most of the density of $N(\mu_i,I)$ is contained within $R_i$:

\begin{align*}
    & \Pr_{X \sim N(\mu_i,I)}[X \in R_i] \ge 1 - \sum_{j \ne i} \Pr_{X \sim N(\mu_i,I)}\left[ [N(\mu_i,I)](x) \le [N(\mu_i,I)](x) \right] \\
    & = 1 - \sum_{j \ne i} \Pr_{Z \sim N(0,1)}[z \ge \|\mu_i - \mu_j \|]  \ge 1 - \left\lceil \frac{4}{\alpha}\right\rceil \cdot \Pr_{Z \sim N(0,1)}\left[z \ge 10 \sqrt{\log(1/\alpha)}\right] \\
    & \ge 1 - \left\lceil \frac{4}{\alpha}\right\rceil \cdot e^{-50 \log(1/\alpha)} \ge \frac{1}{2}
\end{align*}

After this observation, we may conclude by definition of $\alpha$-consistency,

\begin{equation*}
    1 \ge \sum_{i=1}^{l} \Pr_{X \sim D}[X \in R_i] \ge \sum_{i=1}^{l} \alpha \cdot \Pr_{X \sim N(\mu_i,I)}[X \in R_i] \ge l \cdot \alpha \cdot \frac{1}{2} = 2,
\end{equation*}
which is a contradiction. Let us define $q_i \triangleq \Pr_{X \sim N(\mu_i,I)}[X \in R_i]$. In summary, we have shown that a list of too many well-separated means that are $\alpha$-consistent would cause the sum of $q_i$ to be large enough to cause a contradiction. 

This proof sketch is enough to easily show the information-theoretic existence of a small list. Let us construct the list iteratively, starting with an empty list $L = []$. At each step, if there is an $\alpha$-consistent $\mu$ that it is at least $10 \sqrt{\log(1/\alpha)}$ away from everything in $L$, then arbitrarily add any such $\mu$. Otherwise, our list is finished. By our above proof, this will terminate in less than $\frac{4}{\alpha}$ steps, and hence yield a list of size $\le \frac{4}{\alpha}$. 
\end{proof}

\subsection{Our result: learning up to accuracy $\eps$}

We will now detail how to get a list with better accuracy $\eps$. First, we may use the previous warm-up as a subroutine, to give us a preliminary list $L_0$ where $|L_0| \le \frac{4}{\alpha}$, and any $\alpha$-consistent $\mu$ will be within $10 \sqrt{\log(1/\alpha)}$ distance of some element in $L_0$.

This will let us consider separately learning within balls of radius $10 \sqrt{\log(1/\alpha)}$. Focusing within these balls may not immediately seem helpful, but will prove useful later. 

Recall $R_i = \{ \left\|x-\mu_i\right\|^2 < \left\|x-\mu_j\right\|^2, \forall j \neq i \}$ and $q_i = \Pr_{x\sim N(\mu_i,I)}[ X \in R_i]$. Our analysis will follow a similar program to the warm-up but with sharper guarantees: we will closely analyze the $q_i$, and then similarly conclude the sum of $q_i$ must exceed $1/\alpha$ if there are too many $\alpha$-consistent $\mu$ that are $\beta$-separated. (Using $\beta = \eps$ will clearly give some helpful guarantee, but later we will get sharper results by using this for general $\beta$.) First, we lower bound the sum of $q_i$:
\begin{theorem}\label{thm:q-bound}
    Assume $\mu_1 \dots, \mu_l$ are $\beta$-separated and $\left\|\mu_i\right\| \le r$, where $l \ge 2^{C r^4 / \beta^6}$ for sufficiently large $C>0$, $0 < \alpha \le \frac{1}{2}$, and $0 < \beta \le C \sqrt{\log(1/\alpha)}$. Then,
    \begin{equation}
        \mathrm{poly}(1/\alpha) \cdot \sum_{i=1}^L q_i \ge 2^{\frac{c\beta^3}{r^2}\sqrt{\log l}}.
    \end{equation}
\end{theorem}

Deferring the proof of \cref{thm:q-bound} to \cref{subsub:q-bound}, let us first see how this implies \cref{thm:info-theory}.

\textbf{Concluding \cref{thm:info-theory}.} First, use \cref{lemma:classic} to obtain a list $L_0$ of size at most $\frac{4}{\alpha}$ such that any $\alpha$-consistent $\mu$ is within $r_0 \triangleq 10\sqrt{\log(1/\alpha)}$ of some item in $L_0$.

For each $\mu' \in L_0$, we will create a sub-list $L'$. Without loss of generality, translate the distribution so $\mu'$ is the origin. We now aim for a list with error at most $\eps$. Our goal will be complete if for any $\alpha$-consistent $\mu$ where $\|\mu - \mu' \| \le r_0$, there is a $\mu^* \in L'$ where $\|\mu - \mu^*\| \le \eps$. 

A naive application of \cref{thm:q-bound} would give some upper bound on the size of each list $\mu_1,\dots,\mu_l$ by setting $r = 10 \sqrt{\log(1/\alpha)}$ and $\beta = \eps$. This would yield a list upper bound of size $\exp(C \log^4(1/\alpha)/\eps^6)$. However, we may apply \cref{thm:q-bound} more carefully and do better.

Consider the union of all sub-lists to be $\mu_1,\dots,\mu_l$, where all $\mu_i$ are $\alpha$-consistent and $\eps$-separated. Let $|M_i|$ denote the size of the smallest $\frac{r_0}{2^i}$-cover of the list $\mu_1,\dots,\mu_l$. By construction, $|M_0| \le 4/\alpha$.

\begin{claim}\label{claim:m-bound}
    For any positive integer $i$, it holds that $|M_i| \le |M_{i-1}| \cdot \exp(C\log^2(1/\alpha)/r_i^2)$.
\end{claim}
\begin{proof}
    For constructing the $r_i\triangleq \frac{r_0}{2^i}$-cover, we may separately consider covering the list entries within the $r_{i-1}$-radius balls centered at the cover centers $M_{i-1}$. The covering number within each $r_{i-1}$-ball is at most the size of the largest set of $r_{i}$-separated $\mu_j$ in this ball. Suppose there were a collection of $l'$ items $\mu_j$ in our list that are $r_i$-separated in this $r_{i-1}$-radius ball. Then, using \cref{thm:q-bound} with $r=r_{i-1}$ and $\beta = r_i$, we conclude:
    \begin{align*}
        \sum_{i=1}^{l'} \alpha \Pr_{X \sim D}[X \in R_i] \le 1 &\iff \sum_{i=1}^{l'} q_i \le 1/\alpha\\
         &\implies 2^{\frac{c r_i^3}{r_{i-1}^2} \sqrt{\log l'}} \cdot  \poly(\alpha) \le 1 \\
        & \iff 2^{r_i \sqrt{\log l'}} \cdot  \poly(\alpha) \le 1 \\
        & \implies l' \le \exp(C\log^2(1/\alpha)/r_i^2) \quad\qedhere
    \end{align*}
\end{proof}
Let $i^* \triangleq \lceil \log_2(\frac{r_0}{\eps/4})\rceil$. Observe how it must hold that $l = |M_{i^*}|$, since the list $\mu_1,\dots,\mu_l$ is $\eps$-separated and $r_{i^*}<\eps/2$. Using \cref{claim:m-bound}, we may conclude our proof by computing
\begin{align*}
    l &= |M_{i^*}| = |M_0| \cdot \Pi_{i=1}^{i^*} \frac{|M_i|}{|M_{i-1}|}\\
    & \le \frac{4}{\alpha} \cdot \Pi_{i=1}^{i^*} \exp(C\log^2(1/\alpha)/r_i^2) \\
    & \le \exp(C\log^2(1/\alpha)/\eps^2).
\end{align*}
This completes the proof of our information-theoretic result.

\textbf{Remark for finite samples. } Let us roughly discuss how this proof technique would yield results for the typical list-decodable setting. (We only discuss roughly since the later result \cref{thm:efficient} will directly prove this.) Observe how the above proof mostly leverages that if some $\mu'$ is $\alpha$-consistent, then $\Pr_{X \sim D}[X \in R_i] \ge \alpha \Pr_{X \sim N(\mu',I)}[X \in R_i]$. Imagine if we defined a similar notion of being \textit{finitely-$(\alpha,\eta,l)$-consistent}: where a $\mu'$ is finitely-$(\alpha,\eta,l)$-consistent on a normalized empirical distribution $D'$ if $\Pr_{X \sim D'}[X \in H] \ge \alpha \Pr_{X \sim N(\mu_i,I)}[X \in H] - \eta$, for all $H$ that are the intersection of at most $l$ halfspaces. Since the intersection of $l$ halfspaces is a concept class with bounded VC dimension, we may show that the true $\mu$ will be finitely-$(\alpha,\eta,l)$-consistent for a sufficiently large $n$ with high probability. Moreover, this condition is enough for our proof to work: we could follow a similar proof structure as before, repeatedly choosing any $\mu'$ that is $\eps$-separated from the existing list and is finitely-$(\alpha,\eta,l)$-consistent, until there is no such $\mu'$.

Employing this plan efficiently poses numerous challenges that we approach in \cref{sec:efficient}.

\subsubsection{Proof of \cref{thm:q-bound}}\label{subsub:q-bound}
\begin{proof}

We now prove the main theorem for our identifiability result. We consider the following auxiliary probability sequence $(p_i)_{i\in [l]}$, defined as $p_i = \Pr_{x\sim N(0,I)}[x\in A_i]$, where $A_i = \{x: \langle x, \mu_i\rangle > \langle x,\mu_j \rangle, \forall j\neq i\}$. Let $R'_i$ correspond to the region $R_i$ translated by $-\mu_i$. By definition,

\begin{equation*}
    q_i = \Pr_{X \sim N(\mu_i,I)}[X \in R_i] = \Pr_{X \sim N(0,I)}[X \in R_i'] = \Pr_{X \sim N(0,I)}[\|x\|^2 < \|x - \mu_j + \mu_i\|, \forall j \ne i]. 
\end{equation*}

We initially observe that $q_i \ge p_i$ via the following simple analysis that $R_i' \supseteq A_i$:
\begin{equation*}
    R_i' = \{x : \left\|x\right\|^2 < \left\|x+(\mu_i-\mu_j)\right\|^2, \forall j \neq i\} \supseteq \{ x : \langle x, \mu_i \rangle > \langle x, \mu_j \rangle, \forall j \neq i\} = A_i
\end{equation*}

More consequentially, we may show $R_i'$ actually contains an \textit{fattening} of $A_i$; combined with the Gaussian isoperimetry inequality, this will give us a favorable lower bound for $q_i$ in terms of $p_i$:
\begin{lemma}\label{lemma:p-bound-q}
    $q_i \ge \alpha^{c'} \cdot p_i \cdot \exp(c\beta \sqrt{\log(1/p_i)}).$
\end{lemma}
\begin{proof}
    We show that $R_i'$ contains a fattening of $A_i$:
    \begin{claim}\label{claim:q-fat}
        $(A_i)_{\beta/2} \triangleq  \{x: \mathrm{dist}(x,A_i)\le \beta\} \subseteq \{x: \left\|x\right\|^2 < \left\|x-\mu_j+\mu_i\right\|^2, \forall j\neq i\} = R_i'$
    \end{claim}
    \begin{proof}
    For $y\in (A_i)_{\beta/2}$, write it as $x+h$, where $x\in A_i$ and $\left\|h\right\| \le \beta/2$. Then for any $j\neq i$
    \[
        \langle y, \mu_j - \mu_i \rangle = \langle x, \mu_j - \mu_i \rangle + \langle h, \mu_j - \mu_i \rangle < \langle h, \mu_j - \mu_i \rangle \le \left\|h\right\| \left\|\mu_i - \mu_j\right\| \le \frac12 \left\|\mu_j - \mu_i\right\|^2,
    \]which implies $\left\|y\right\|^2 < \left\|y-\mu_j+\mu_i\right\|^2$ as we desired.
    \end{proof}

    Since $\Pr_{x\in N(0,I)}[x\in A_i] = p_i$, we will use the Gaussian isoperimetry inequality to get a lower bound for the measure of the fattening $\Pr_{x\in N(0,I)}[x\in (A_i)_{\beta/2}]$. We use the following implication of the Gaussian isoperimetry inequality and defer the proof to \cref{app:isoper}:

    \begin{lemma}[Gaussian isoperimetry]\label{lemma:isoperimetry}
    For a measurable set $A$ in $\mathbb{R}^n$ under standard gaussian measure $\mu$, let $A_\beta = \{x\in \mathbb{R}^n: \mathrm{dist}(x, A) \le \beta\}$. If $\mu(A) \le 1/2$, then
    \[
        \mu(A_\beta) \ge \mu(A) \cdot \exp\left(-c'(1+\beta^2)+c\beta\sqrt{\log(1/\mu(A))}\right).
    \]
\end{lemma}

    Using \cref{claim:q-fat}, \cref{lemma:isoperimetry}, $\beta \le  O(\sqrt{\log(1/\alpha)})$, and that $p_i \le \frac{1}{2}$ by symmetry, we conclude 
    
    \[
        q_i \ge \Pr_{x\in N(0,I)} [x\in (A_i)_{\beta/2}] \ge \alpha^{c'} \cdot p_i \cdot \exp\left(c\beta \sqrt{\log(1/p_i)}\right),
    \]which completes the proof of our lemma.
\end{proof}

We now hope to use \cref{lemma:p-bound-q} to prove \cref{thm:q-bound}. We remark that if $p_1,\dots,p_l$ could be any arbitrary non-negative values that sum to $1$, then our approach would not yield \cref{thm:q-bound}. For example, if $\beta=1$ and our vector $p$ was the infinite-length vector $\frac{1}{2},\frac{1}{4},\dots$, then the quantity $\sum_{i} p_i \cdot \exp(c \beta \sqrt{\log(1/p_i)})$ from \cref{lemma:p-bound-q} would converge to a constant, yet our desired bound for \cref{thm:q-bound} will go to infinity as $l \rightarrow \infty$. 

Luckily for our approach, the values of $p_i$ are not arbitrary. Each $p_i$ corresponds to probability that $\mu_i$ is the maximizer for a Gaussian process $\max_i \langle x, \mu_i\rangle$ for $X \sim N(0,I)$. We will prove this distribution of maximizers is constrained in a way that enforces the distribution $p$ must take values such that the quantity $\sum_{i} p_i \cdot \exp(c \beta \sqrt{\log(1/p_i)})$ grows according to our desired theorem.

\textbf{Warm-up intuition: infinite dyadic $p = [\nicefrac{1}{2},\nicefrac{1}{4},\dots]$ is impossible. } As a warm-up, let us quickly show why it is impossible for the vector $p$ to take value $[\nicefrac{1}{2},\nicefrac{1}{4},\dots]$. By Sudakov's minoration inequality (\cref{thm:sudakov-minoration}), we know the expected maximum $\Ex[\sup_{i \in [l]} \langle x, \mu_i \rangle] \ge c \beta \sqrt{\log(l)}$ is infinite in this example where $l$ is infinite. However, if $p$ is dyadic and corresponds to the distribution of maximizers, then we may conclude this expected maximum is finite:
\begin{align*}
    & \Ex_{X \sim N(0,I)}[\sup_{i\in [l]} \langle x, \mu_i \rangle ] = \sum_{i=1}^{\infty} \Ex_{X \sim N(0,I)}[\mathbbm{1}_{\mu_i \textrm{ maximizes}} \cdot \langle x, \mu_t \rangle ] \\
    & \le \sum_{i=1}^{\infty} \sqrt{\Ex_{X \sim N(0,I)}[\mathbbm{1}_{\mu_i \textrm{ maximizes}}^2] \cdot \Ex_{X \sim N(0,I)}[\langle x, \mu_i \rangle^2] } \tag{via Cauchy–Schwarz} \\
    & = \sum_{i=1}^{\infty} \sqrt{2^{-i} \cdot \| \mu_i \|^2} \le Cr
\end{align*}

This contradicts the lower bound from Sudakov's minoration inequality, and shows that $p$ could not possibly take the value of this infinite dyadic distribution. In general, this gives some sense that the distribution of $p$ must be well-spread in a way that hopefully gives us a favorable lower bound for $\sum_{i} p_i \cdot \exp(c \beta \sqrt{\log(1/p_i)})$. In the remaining proof, we will judiciously apply Holder's inequality in a way that attains our desired result.

\textbf{General proof.} By \cref{lemma:p-bound-q}, we know
\[
    \sum_{i\in [l]}q_i \ge \alpha^{c'} \cdot \sum_{i\in [l]} p_i \cdot \exp(\beta \sqrt{\log(1/p_i)}).
\]

We use Holder's inequality on the expectation of the Gaussian process, using positive constants $s, t$ that satisfy $\frac{1}{s}+\frac{1}{t}=1$ (we choose $s,t$ later),
\[
    \sum_{i\in [l]} \Ex[\mathbbm{1}_{x\in A_i} \cdot \langle x, \mu_t \rangle ] \le \sum_{i\in [l]}(\Ex[\mathbbm{1}_{x\in A_i}])^{1/t} \cdot \Ex[|\langle x, \mu_t \rangle|^{s} ]^{1/s} = \sum_{i\in [l]} (p_i)^{1-1/s} \cdot O\left(r\sqrt{s}\right).
\]

By Sudakov’s minoration inequality (\cref{thm:sudakov-minoration}) we know the left-hand side is at least $c\beta \sqrt{\log l}$. We will choose $s = \frac{c^2 \beta^2 \log l}{9r^2}$, which satisfies $s > 1$ by our condition on $l$. Using this choice of $s$:
\begin{equation*}
    c \beta \sqrt{\log l} \le \sum_{i \in [l]}(p_i)^{1-1/s} \cdot O(r \sqrt{s}) \implies \sum_{i\in [l]} (p_i)^{1-1/s} \ge 3.
\end{equation*}

We observe this sum is still large even when restricting to $S = \{i\in [l]: p_i\ge 1/l^2\}$:
\begin{align*}
    &\sum_{i\in S}(p_i)^{1-1/s} = \sum_{i\in [l]}(p_i)^{1-1/s} - \sum_{i\notin S}(p_i)^{1-1/s} \ge  3 - \sum_{i\notin S}(p_i)^{1-1/s} \\
    & \ge 3 - \sum_{i\notin S} (p_i)^{1/2}  \ge 3 - \sum_{i\notin S} 1/l \ge 2 \tag{using $s \ge 2$ from our condition on $l$}
\end{align*}

Further, for $i\in S$, we have $(p_i)^{1-1/s} \le p_i\cdot \exp\left(\frac{18r^2\sqrt{\log(1/p_i)}}{c^2\beta^2\sqrt{\log l}}\right)$, this is because
\begin{align*}
    & \qquad (p_i)^{1-1/s} \le p_i\cdot \exp\left(\frac{18r^2\sqrt{\log(1/p_i)}}{c^2\beta^2\sqrt{\log l}}\right)\\
    & \impliedby (1/p_i)^{1/s} \le \exp\left(\frac{18r^2\sqrt{\log(1/p_i)}}{c^2\beta^2\sqrt{\log l}}\right)\\
    & \impliedby \frac{9r^2}{c^2\beta^2 \log{l}} \log(1/p_i) \le \frac{18r^2\sqrt{\log(1/p_i)}}{c^2\beta^2\sqrt{\log l}} \\
    & \impliedby \sqrt{\log(1/p_i)} \le 2\sqrt{\log l} \\
    & \impliedby p_i \ge 1/l^4.
\end{align*}

Thus,
\[
    \sum_{i\in [l]} p_i\cdot \exp\left(\frac{18r^2\sqrt{\log(1/p_i)}}{c^2\beta^2\sqrt{\log l}}\right) \ge \sum_{i\in S} p_i\cdot \exp\left(\frac{18r^2\sqrt{\log(1/p_i)}}{c^2\beta^2\sqrt{\log l}}\right) \ge \sum_{i\in S}p_i^{1-1/s} \ge 2.\numberthis \label{step:ge2}
\]

With this, we may conclude the proof of our theorem:
\begin{align*}
    & \alpha^{-c'} \cdot\sum_{i \in [l]} q_i \ge \sum_{i\in [l]} p_i \cdot \exp(\beta \sqrt{\log(1/p_i)}) \\
    & \ge \left(\sum_{i\in [l]} p_i\cdot \exp\left(\frac{18r^2\sqrt{\log(1/p_i)}}{c^2\beta^2\sqrt{\log l}}\right)\right)^{\frac{c^2\beta^3}{18r^2}\sqrt{\log l}} \tag{Jensen's inequality; $l$ chosen s.t. $\frac{c^2 \beta^3}{18r^2} \sqrt{\log l} \ge 1$} \\
    & \ge 2^{\frac{c^2\beta^3}{18r^2}\sqrt{\log l}}. \tag{via \cref{step:ge2}} \quad\qedhere
\end{align*}

\end{proof}

\subsection{List size lower bound}
Theorem 1.5 of  \cite{diakonikolas2018list} also shows how it is information-theoretically impossible to get better than $\Omega(\sqrt{\log1/\alpha})$ accuracy with a list of only size $\poly(1/\alpha)$. For a general lower bound in terms of the list size required for $\eps$ accuracy, the same construction of \cite{diakonikolas2018list} will yield a lower bound qualitatively similar to our upper bound in the previous section. 
Interestingly, we note that for list-decodable learning up to accuracy $\eps$, list size of $2^{\Omega(\log^2(1/\alpha)/\eps^2)}$ was necessary. Observe how when $\eps = \Theta(\sqrt{\log (1/\alpha)})$, this size is $\poly(1/\alpha)$, as expected. The proof for this lower bound follows the proof strategy of Proposition 5.11 in \cite{diakonikolas2023algorithmic}; we present this for completeness in \cref{app:lb}.

\begin{lemma}\label{lemma:lb}
    Let $0 < \eps < \frac{\sqrt{\log(\nicefrac{1}{2\alpha})}}{2}$ and $0 < \alpha < \nicefrac{1}{3}$. There exists a distribution $D$ such that for any list containing an $\eps$-close mean for all possible $\alpha$-consistent means, the list must have size at least $\exp(c \log^2(1/\alpha)/\eps^2)$.
\end{lemma}

\section{Efficient list-decodable learning}\label{sec:efficient}
In this section, we will design an algorithm that efficiently constructs a list of size comparable to our information-theoretic upper bound of \cref{thm:info-theory}, given access to a finite number of samples. We gave an overview of the proof in \cref{subsec:efficient-overview}, and formally state our result here:
\begin{theorem}\label{thm:efficient}
    In the setting of Definition~\ref{def:main}, there is an algorithm which outputs a list of at most
    \[
    L = \exp \left(O \left( \frac{\log^2 (1 / \alpha)}{\eps^2} \right) \right)
    \]
    candidate means, which achieves error $\eps$ with probability at least $0.99$ when $n \ge d^{C\log^2 (1/\alpha)/\eps^2} + \exp(L)$. The time complexity is 
    \[
    d^{C\log^2 (1/\alpha)/\eps^2} + \exp \left(\exp \left( \widetilde{O} \left( \frac{\log^2(1/\alpha)} {\eps^2} \right) \right) \right) \; .
    \]
\end{theorem}

In \cref{subsec:subspace}, we will find a low-dimensional subspace that is close to $\mu$ with high probability. In \cref{subsec:exhaust}, we focus on this low-dimensional subspace and exhaustively search over a net for points that approximately match moments as desired. In \cref{subsec:fool}, we will detail the required technical result for how approximately matching moments fools halfspaces. In \cref{subsec:combine}, we finally combine these ingredients to conclude our algorithm.

\input{subspace}

\subsection{Exhaustive search for moment-matching candidates}\label{subsec:exhaust}
\input{exhaustive}

\subsection{Sharper moment-matching fooling}\label{subsec:fool}
\input{fool-moments}

\subsection{Combining ingredients: proof of \cref{thm:efficient}}\label{subsec:combine}
\input{combining}

\bibliography{ref}

\appendix

\input{technical-lemmas}
\end{document}

%% file: macros.tex
\documentclass[11pt,letterpaper]{article}%
\usepackage[utf8]{inputenc}
\usepackage[rflt]{floatflt}
\usepackage{fullpage}
\usepackage[T1]{fontenc}
\usepackage{graphicx}
\usepackage{epsfig}
\usepackage{color}
\usepackage{enumerate}
\usepackage{verbatim}
\usepackage{amsthm,amssymb}
\usepackage{amsmath}
\usepackage{thmtools,thm-restate}
\usepackage{wrapfig}
\usepackage{xspace}
\usepackage{amsfonts}
\usepackage{bbm}
\usepackage{algorithm, algpseudocode, caption, subcaption}

\usepackage{todonotes}

\definecolor{darkgreen}{rgb}{0,0.5,0}
\usepackage{hyperref}
\hypersetup{
    unicode=false,          %
    colorlinks=true,        %
    linkcolor=blue,          %
    citecolor=darkgreen,        %
    filecolor=magenta,      %
    urlcolor=cyan           %
}

\usepackage[disableredefinitions,roman]{complexity}
\usepackage[framemethod=tikz]{mdframed}
\global\mdfdefinestyle{myframe}{leftmargin=.75in,rightmargin=.75in,linecolor=black,linewidth=1.5pt,innertopmargin=10pt,innerbottommargin=10pt} 
\usepackage{mdframed}
\usepackage{framed}
\usepackage{nicefrac}

\date{}

\newtheorem{theorem}{Theorem}[section]
\newtheorem{lemma}[theorem]{Lemma}
\newtheorem{claim}[theorem]{Claim}

\newtheorem{definition}[theorem]{Definition}
\newtheorem{corollary}[theorem]{Corollary}

\usepackage[capitalize, nameinlink]{cleveref}
\crefname{theorem}{Theorem}{Theorems}
\Crefname{lemma}{Lemma}{Lemmas}
\Crefname{alg}{Algorithm}{Algorithms}
\Crefname{claim}{Claim}{Claims}
\Crefname{infclaim}{Claim}{Claims}
\Crefname{observation}{Observation}{Observations}
\Crefname{invariant}{Invariant}{Invariants}
\Crefname{algorithm}{Algorithm}{Algorithms}

\newcommand{\eps}{\varepsilon}

\newcommand{\bbR}{\mathbb{R}}

\DeclareMathOperator*{\Ex}{\mathbb{E}}

\newcommand\numberthis{\addtocounter{equation}{1}\tag{\theequation}}
\newcommand{\norm}[1]{\left\| #1 \right\|}

\newcommand{\Sym}{\mathrm{Sym}}
\renewcommand{\flat}{\mathrm{flat}}
\renewcommand{\Pr}{\operatorname{{Pr}}}
\DeclareMathOperator*{\argmax}{arg\,max}

\allowdisplaybreaks[1]

\newcommand{\Sgood}{S_{\mathrm{good}}}
\newcommand{\iprod}[1]{\left\langle #1 \right\rangle}

%% file: subspace.tex
\subsection{Subspace learning}\label{subsec:subspace}\label{sec:subspace}

In this section, we give the first key algorithmic subroutine we will need for our efficient algorithm, namely, a method which allows us to recover a low-dimensional subspace which approximately contains all of the candidate means.
More formally, we show:

\begin{theorem}\label{thm:subspace-learning}
    Let $\alpha, \eps > 0$.
    Let $S$ be an $\alpha$-pure dataset with respect to $N(\mu, I)$ of size 
    \[
    n = \left(\frac{d\log(1/\alpha)}{\eps}\right)^{\Omega(\log^2(1/\alpha)/\eps^2)} \; ,
    \]
    and assume that $\norm{\mu}_2 \leq C \sqrt{\log (1 / \alpha)}$.
    There is an algorithm which, given $S$, outputs a subspace $V \subset \mathbb{R}^d$ of dimensionality $k^{O(k)}$, where $k = 100 \log^2(1/\alpha)/\eps^2$, so that with probability at least $0.99$, we have that $\norm{\Pi_{V^\perp} \mu}_2 \leq \eps$.
    Moreover, this algorithm runs in time $\poly(n)$.
\end{theorem}

We begin with preliminaries, then give an overview, and finally detail our analysis.

\input{tensors}

\subsubsection{Overview}
\paragraph{Overview of Algorithm} 
Let $k = 100 \log^2(1/\alpha)/\eps^2$, $\gamma = e^{-100k}$ and $\lambda = e^{-10k}$.
Recall that $H_t$ denotes the $t$-th Hermite polynomial tensor. 
For any set $T \subseteq S$, we define 
    \[
        {M_t}(T) := \frac{1}{|T|}\sum_{i\in T} H_t(X_i) \; .
    \]
Our algorithm proceeds in two steps.
First, we perform an iterative filtering algorithm to obtain a subset $T$ of $S$ with the guarantee that (i) we have removed essentially no good points from $T$, and (ii) $\norm{M_t (T)}_2 \leq \frac{1}{\gamma}(C\sqrt{\log(1/\alpha)})^{t}$, for all $t \in [2k]$.

Once we have done this, we show that a simple spectral method applied on a flattening of the moment tensors allows us to extract the desired subspace.
Formally, observe that every $M_t$ can be written naturally as $(d \times d^{t - 1})$-sized matrix by flattening the first tensor mode, and denote this matrix $\mathrm{flat} (M_t(T))$.
Our algorithm will simply take the top left singular vectors of $\mathrm{\flat} (M_t (T))$ for all $t = 1, \ldots, 2k$, and take the span of their union.
We provide formal pseudocode for this algorithm in Algorithm~\ref{alg:subspace}. In the remaining proof, may use $M_t$ as shorthand for $M_t(T)$.

\begin{algorithm}
\caption{Subspace learning algorithm}
\label{alg:subspace}
\begin{algorithmic}
   \State Let $T = [n]$.
   \While {there exists a $t \in \{1,2,...,2k\}$ such that $\left\|{M_t}(T)\right\|_2 > \gamma^{-3} t! (C\sqrt{\log(1/\alpha)})^{t}$}
    \State For every $i \in T$, let 
    \[
        \tau_i = \max\left(0,  \langle H_t(X_i), {M_t}(T) \rangle + \gamma^{-2} t! \left(C \sqrt{\log 1 / \alpha} \right)^t \cdot \norm{M_t}_2  \right) \; .
    \]
    \State Sample $i \propto \tau_i$, and remove $i$ from $T$. 
   \EndWhile

   \For {$t=1,2...2k$}
   \State Let $Q_t$ denote the set of the $s_t$ left singular vectors of $\flat (M_t (T))$, where
   \[
    s_t = \gamma^{-6} (t!)^2 (C\sqrt{\log(1/\alpha)})^{2t} / \lambda^2 \; .
   \]
   \EndFor
   \State Return
    \[
        V = \mathrm{span} \left( \bigcup_{i = 1}^{2k} Q_t \right) \; .
    \]
\end{algorithmic}
\end{algorithm}

\paragraph{Regularity condition} Before we analyze the algorithm, it will be convenient to establish the following regularity condition on the good set of points.
Recall that by assumption, we know that there exists a set $\Sgood \subset S$ so that $|\Sgood| = \alpha n$, and $\Sgood$ is a set of independent draws from $N(\mu, I)$.
We will require the following lemma, which follows by \cref{lem:hermite-concentration}, \cref{lem:hermite-tensor-upper} (for \cref{eqn:hermite-score-second-moment}) and \cref{lem:gaussian-concentration} (for \cref{eqn:gauss-2-concent,eqn:gauss-4-concent}).

\begin{lemma}
    \label{lem:hermite-concentration-2}
    Let $X_1, \ldots, X_m \sim N(\mu, I)$, and let $k$ be a positive integer.
    Then, for 
    \[
    n \geq (dk\sqrt{\log(1/\alpha)})^{\Omega(k)} \; ,
    \]
    with probability $0.99$ the following holds: 
    for all $t \leq 2k$, and for all $u \in \bbR^{d^t}$, we have that
    \begin{equation}\label{eqn:hermite-score-second-moment}
        \frac{1}{m} \sum_{i=1}^m \langle u, H_k(X_i) \rangle^2 \leq t! \cdot C^t \cdot \exp \left( \sqrt{t} \norm{\mu}_2 \right) \cdot \norm{u}_2^2 \; ,
    \end{equation}
    for some universal constant $C$, and for all unit vector $w$, we have that
    
    \begin{equation}\label{eqn:gauss-4-concent}
        \frac{1}{m} \sum_{i=1}^m h_k^4(\langle X_i, w \rangle) \le \Ex_{X\sim N(\mu, I)} [h_k^4(\langle X, w \rangle)] + O(1)  
    \end{equation}
    
    \begin{equation}\label{eqn:gauss-2-concent}
        \frac{1}{m} \sum_{i=1}^m h_k^2(\langle X_i, w \rangle) \ge \Ex_{X\sim N(\mu, I)} [h_k^2(\langle X, w \rangle)] - O(1)
    \end{equation}
\end{lemma}

\noindent
For the rest of the proof, condition on the event that Lemma~\ref{lem:hermite-concentration-2} holds for $\Sgood$.
By our condition on $n$, this happens with probability at least $0.999$.

\subsubsection{Analysis of the filter}

First, observe that the filtering routine clearly runs in time which is polynomial in $n$, since each loop runs in time $\poly (n, d^k) = \poly (n)$, and we will run for at most $n$ loops, since we remove one data point each time we run the filter.

Our main result for the filtering subroutine is the following:
\begin{lemma}
\label{lem:filter-main}
    For the output set $T$, we have $\left\|{M_t}(T)\right\|_2 \le \frac{1}{\gamma}(C\sqrt{\log(1/\alpha)})^{t}$ and with probability $1-\delta$, we have $|S_{\mathrm{good}} \cap T| \ge (1-\delta)|S_{\mathrm{good}}|$, where $\delta = \frac{\gamma^{1/2}}{\alpha} = e^{-\Omega(k)}$. 
\end{lemma}
\begin{proof}
    By the termination condition of the filter, we have that $\left\|{M_t}(T)\right\|_2 \le \frac{1}{\gamma}(C\sqrt{\log(1/\alpha)})^{t}$. 

    Thus, it remains to prove the second statement.
    We first observe that for any iteration for $t$, we have
    \begin{align*}
    \frac{1}{|T|} \sum_{i \in T} \tau_i &\geq \frac{1}{|T|} \sum_{i \in T}  \langle H_t(X_i), {M_t}(T) \rangle \\
    &= \norm{M_t (T)}_2^2 \; . 
    \end{align*}
    On other other hand, we claim that
    \begin{align}
        \label{eq:good-score}
        \frac{1}{|T|} \sum_{i \in \Sgood} \tau_i \leq \gamma \norm{M_t (T)}_2^2 \; ,  
    \end{align}
    where for simplicity, we say that $\tau_i = 0$ for any $i$ that has already been removed from the dataset.
    Given this, we have that 
    \[
    \frac{\sum_{i\in S_{\mathrm{good}}} \tau_i }{\sum_{i\in T} \tau_i} \leq \gamma \; ,
    \]
    which implies that if we define the potential function
    \[
        \Phi := |T| + \frac{1}{\gamma}(|S_{\mathrm{good}}| - |S_{\mathrm{good}}\cap T|) \; ,
    \]
    then this potential function is a super-martingale.
    Therefore, with probability $1 - \gamma^{1/2}$, at termination, $\Phi \leq \gamma^{-1/2} n$, which implies that $|S_{\mathrm{good}}\cap T| \geq (1-\frac{\gamma^{1/2}}{\alpha} )|S_{\mathrm{good}}|$, as claimed.
    Thus it suffices to prove~\eqref{eq:good-score}.

    First, by Jensen's inequality, we observe that 
    \begin{align*}
    \left| \frac{1}{|\Sgood|} \sum_{i \in \Sgood} \langle M_t, H_t (X_i) \rangle \right| &\leq \left( \frac{1}{|\Sgood|} \sum_{i \in \Sgood} \langle M_t, H_t (X_i) \rangle^2 \right)^{1/2} \\
    &\leq \left( t! C^t \exp \left( \sqrt{t}  \norm{\mu}_2 \right) \right)^{1/2} \norm{M_t}_2 \\
    &\ll \frac{\gamma}{10} \norm{M_t}_2^2 \; .
    \end{align*}
    Let $A_t$ denote the set of points in $\Sgood$ satisfying $\langle M_t, H_t (X_i) \rangle \leq - \gamma^{-2} t! \left(C \sqrt{\log 1 / \alpha} \right)^t \norm{M_t}_2$.
    By Chebyshev's inequality with \cref{lem:hermite-concentration-2},
    \begin{align*}
        \frac{|A_t|}{|\Sgood|} &\leq \frac{ t! C^t \exp \left( \sqrt{t}  \norm{\mu}_2 \right) \cdot \norm{M_t}_2^2}{\left(\gamma^{-2} t! (C \sqrt{\log 1 / \alpha} )^t\right)^2 \norm{M_t}_2^2} \\
        &\ll \gamma^{3} \; ,
    \end{align*}
    by our choice of $\gamma$.
    In particular, this implies that
    \begin{align*}
    \left| \frac{1}{|\Sgood|} \sum_{i \in A_t} \langle M_t, H_t (X_i) \rangle  \right| &\leq \gamma^{1.5} \cdot \left( t! C^t \exp \left( \sqrt{t}  \norm{\mu}_2 \right) \right)^{1/2} \norm{M_t}_2 \\
    & \ll \frac{\gamma}{10} \norm{M_t}_2^2 \; .
    \end{align*}

    Thus, overall, we have that
    \begin{align*}
        \frac{1}{\Sgood} \sum_{i \in \Sgood} \tau_i &= \frac{1}{|\Sgood|}\sum_{i \in \Sgood \setminus A_t} \iprod{M_t, H_t (X_i)} + \gamma^{-2} t! (C \sqrt{\log 1 / \alpha})^t \norm{M_t}_2 \\
        & \leq \frac{\gamma}{5} \norm{M_t}_2^2 + \gamma^{-2} t! (C \sqrt{\log 1 / \alpha})^t \norm{M_t}_2 \\
        &\ll \gamma \norm{M_t}_2^2 \; ,
    \end{align*}
    as claimed.
    This completes the proof.
\end{proof}

\subsubsection{Analysis of the spectral truncation}

First, notice that since
$\left\|M_t(T)\right\|_2^2 \le \gamma^{-6} t! (C\sqrt{\log(1/\alpha)})^{2t}$, this implies that the $(s_t + 1)$-th singular value of $\flat (M_t (T))$ is at most $\lambda$. 
By the definition of $Q_t$, for any $1\le t\le 2k$, we have for any $u\in \mathbb{R}^d: u\perp V$ and $H\in \mathbb{R}^{d^{t-1}}$,
\[
    \langle M_t(T), u\otimes H\rangle \le \lambda \left\| u \right\|_2 \left\| H \right\|_2.
\]

We next require the following lemma:
\begin{lemma}\label{lem:upper-bound-test}
    For any unit vector $w\in \mathbb{R}^d$ that satisfies $w\perp V$, we have \[\frac{1}{|T|}\sum_{i\in T} h_k^2(\langle X_i, w \rangle) \le 2\Ex_{x\sim N(0,1)}[h_k^2(x)] = 2\cdot k!.\]
\end{lemma}

\begin{proof}
    We decompose  $h_k^2(x)$ by Hermite polynomials: $h_k^2(x) = \sum_{t=0}^{2k} c_t h_t(x)$, then \[c_0 = \Ex_{x\sim N(0,1)}[h_k^2(x)] = k!.\]
Hence
\begin{align*}
    \frac{1}{|T|}\sum_{i\in T} h_k^2(\langle X_i, w \rangle) & = \frac{1}{|T|} \sum_{i\in T} \left(c_0 + \sum_{t=1}^{2k}c_t h_t(\langle X_i,w \rangle) \right) \\
    & = k! + \sum_{t=1}^{2k}\frac{c_t}{|T|} \sum_{i\in T} h_t(\langle X_i,w \rangle) \\
    & = k! + \sum_{t=1}^{2k}\frac{c_t}{|T|} \sum_{i\in T} \langle H_t(X_i),w^{\otimes t} \rangle \tag{By \cref{claim:hermite}} \\
    & = k! + \sum_{t=1}^{2k} c_t \langle M_t(T),w^{\otimes t} \rangle \\
    & \le k! + \lambda \sum_{t=1}^{2k} c_t. \tag{by $w\perp V$}
\end{align*}
\noindent
We have following bound on $\sum_{t=1}^{2k} c_t$:
\begin{align*}
    \left(\sum_{t=1}^{2k} c_t \right)^2 & \le \sum_{t=1}^{2k} \frac{1}{t!} \cdot \sum_{t=1}^{2k} (t!) c_t^2 \tag{Cauchy-Schwarz} \\
    & \le e \Ex_{x\in N(0,1)}[\big(\sum_{t=1}^{2k}c_t h_t(x)\big)^2] \\
    & \le 3 \Ex_{x\in N(0,1)}[h_k^4(x)] \\ 
    & \le 3 \cdot 3^{2k} (\Ex_{x\in N(0,1)}[h_k^2(x)])^2 = 3^{2k+1} (k!)^2, \tag{By hypercontractivity}
\end{align*}
and so the lemma follows by our choice of $\lambda = e^{-10k}$.
\end{proof}

\noindent 
Now we are ready to prove the main theorem of this section:
\begin{proof}[Proof of \cref{thm:subspace-learning}]
Suppose the distance of $\mu$ to $V$ is larger than $\eps$. 
Let $w$ be the unit vector parallel to $\big(\Pi_{V^{\perp}}\big) \mu$.

On the one hand, by \cref{lem:upper-bound-test}, we know that
\[\frac{1}{|T|}\sum_{i\in T} h_k^2(\langle X_i, w \rangle) \le 2\cdot k! \; .\]
\noindent
On the other hand, we also have that:
\begin{align*}
    \frac{1}{|T|}\sum_{i\in T} h_k^2(\langle X_i, w \rangle) & \ge \frac{\alpha}{|S_{\mathrm{good}}|} \sum_{i\in S_{\mathrm{good}} \cap T} h_k^2(\langle X_i, w \rangle) \\
\end{align*}

Since
\begin{align*}
    \Ex_{i\sim S_{\mathrm{good}}} [\mathbf{1}_{i\notin S_{\mathrm{good}} \cap T} \cdot h_k^2(\langle X_i, w \rangle)] & \le \big(\Ex_{i\sim S_{\mathrm{good}}} [\mathbf{1}_{i\notin S_{\mathrm{good}} \cap T}]\big)^{\frac12} \cdot \big(\Ex_{i\sim S_{\mathrm{good}}} [h_k^4(\langle X_i, w \rangle)]\big)^{\frac12} \tag{Cauchy-Schwarz} \\
    & \le \delta^{1/2} \cdot \big(\Ex_{X\sim N(\mu, I)} [h_k^4(\langle X, w \rangle)] + O(1)\big)^{\frac12} \tag{By \cref{eqn:gauss-4-concent}} \\ 
    & \le \delta^{1/2} \cdot \big(\Ex_{x\sim N(0,1)} h_k^4(x+y)\big)^{\frac12} + O(1)\\
    & \le \delta^{1/2} \cdot 3^{k}~ \Ex_{x\sim N(0,1)} h_k^2(x+y) + O(1) \tag{By hypercontractivity} \\
    & \le O(1),
\end{align*}
\noindent
and so we obtain that
\begin{align*}
    \frac{1}{|S_{\mathrm{good}}|} \sum_{i\in S_{\mathrm{good}} \cap T} h_k^2(\langle X_i, w \rangle) & \ge \Ex_{i\sim S_{\mathrm{good}}} [h_k^2(\langle X_i, w \rangle)] - O(1) \\
    & \ge \Ex_{X\sim N(\mu, I)} [h_k^2(\langle X, w \rangle)] - O(1) \tag{By \cref{eqn:gauss-2-concent}} \\
    & = \Ex_{x\sim N(0,1)} h_k^2(x+\langle \mu, w \rangle ) - O(1)  \\
    & \ge \exp(\Omega(\sqrt{k}\eps)) \cdot k! - O(1)  \ge (3/\alpha) \cdot (k!) \; , \tag{By \cref{lem:lower-bound-test}}
\end{align*}
which implies that
\[
    \frac{1}{|T|}\sum_{i\in T} h_k^2(\langle X_i, w \rangle) \ge 3 \cdot k!,
\]
\noindent
which contradicts to the upper bound $2\cdot (k!)$.
This completes the proof.
\end{proof}

%% file: tensors.tex
\subsubsection{Tensor preliminaries}
We will often have to work with high order tensors, so here we will record some useful notation for them.
\begin{definition}[Tensor indexing]
    Let $t$ be a positive integer.
    For any partition of $[t]$ into sets $S_1, \ldots, S_k$, and any tensors $T_1, \ldots, T_k$ of orders $|S_1|, \ldots, |S_k|$ respectively, we let 
    \[
    T_1^{S_1} \otimes \ldots \otimes T_k^{S_k}
    \]
    to denote the order $t$ tensor obtained by taking the tensor product of $T_1$ in the indices $S_1$ with $T_2$ in the indices $S_2$, etc.
\end{definition}

\begin{definition}[Tensor symmetrization]
    Let $A, B$ be tensors of rank $t_1$ and $t_2$ respectively over $\bbR^d$.
    We let the symmetrization of $A \otimes B$, denoted $\Sym (A \otimes B)$ be given by
    \[
    \Sym (A \otimes B) = \frac{1}{\binom{t_1+t_2}{t_1}}\sum_{\stackrel{[t] = S_1 \sqcup S_2}{|S_1| = t_1, S_2 = t_2}} A^{S_1} \otimes B^{S_2} \; ,
    \]
    that is, we sum over all possible ways of combining $A$ and $B$ into a rank $t_1 + t_2$ tensor.
\end{definition}

We note that we will be sometimes a bit loose, and associate a rank $t$ tensor over $\bbR^d$ with a vector in $\bbR^{d^t}$ in the natural way.
We will also need the following way of associating tensors with a canonical matrix:
\begin{definition}
    Let $t \geq 1$.
    For any order $t$ tensor $T$, we let $\flat(T)$ be the $\mathbb{R}^d \times \mathbb{R}^{d^{t - 1}}$ flattening of $T$ on its first mode.
    That is, $\flat (u \otimes H) = u H^\top$ for any $H \in \mathbb{R}^{d^{t - 1}}$, and we can linearly extend the definition of $\flat$ onto all rank-$t$ tensors.
\end{definition}

\paragraph{Hermite polynomial tensors}
Our algorithm will heavily leverage the structure of the Hermite polynomial tensors, the natural multivariate analog of the univariate Hermite polynomials.

\begin{definition}[The (probabilist's) Hermite polynomial tensor]
\label{def:hermite}
    For all non-negative integers $k$, let $h_k: \mathbb{R} \to \mathbb{R}$ denote the degree-$k$ (probabilist's) Hermite polynomial, defined recursively by $h_0 = 1$, $h_1 (x) = x$, and
    \[
    h_{k + 1} (x) = x h_k (x) - k h_{k - 1} (x) \; .
    \]
    Similarly, 
    The degree-$k$ Hermite polynomial tensor $H_k: \bbR^d \to \left(\bbR^d \right)^{\otimes k}$ is similarly defined recursively by $H_0 = 1$, $H_1(x) = x$, and
    \begin{equation}
    \label{eq:hermite}
    H_{k + 1} (x) =  \Sym (x \otimes H_k (x)) - k \Sym (I \otimes H_{k - 1} (x)) \; .
    \end{equation}
\end{definition}
\noindent
The Hermite polynomials have several important properties, some of which we record below:
\begin{claim}[Properties of Hermite polynomials]
\label{claim:hermite}
    Fix any integer $k \geq 0$. Then, we have the following:
    \begin{itemize}
    \item For any $x, y \in \mathbb{R}$, we have that
    \[
        h_k(x+y) = \sum_{j=0}^k \binom{k}{j} y^{k-j} h_j(x) \; .    
    \]
    \item For any $v \in \mathbb{R}^d$, we have that $\langle H_k (x), v^{\otimes k} \rangle = h_k (\langle v, x \rangle)$.
    \item Let $\mu \in \mathbb{R}$. Then $\Ex_{X \sim N(\mu, 1)} [h_k (X)] = \mu^k$.
    Similarly, for $\mu \in \mathbb{R}^d$, we have that 
    \[
    \Ex_{X \sim N (\mu, I)} [H_k (X)] = \mu^{\otimes k} \; .
    \]
    \item For all $k, \ell$, we have that
    \[
    \Ex_{X \sim N(0, 1)} [h_k (x) h_\ell (x)] = k! \cdot \delta_{k = \ell}  \; .
    \]
    \end{itemize}
\end{claim}
\noindent
The following is a useful identity for the second moment of Hermite polynomial tensor, which is a special case of the formula derived in~\cite{kane2021robust} (see also Claim 9.6 of \cite{liu2022clustering} for this specialization):
\begin{lemma}
\label{lem:hermite-second-moment}
For all $k \geq 0$, and for all $\mu \in \mathbb{R}^d$, we have
    \begin{equation}
    \label{eq:hermite-second-moment}
    \Ex_{X \sim N(\mu,I)}[H_k(X) \otimes H_k(X)] = \sum_{S_1,S_2\subset [k],|S_1|=|S_2|}\sum_{\substack{\text{\emph{Matchings} } P \\ \text{\emph{of} } S_1, S_2}} \bigotimes_{\{a,b\}\in P} I^{(a,k+b)} \bigotimes_{c\notin S_1} \mu^{(c)} \bigotimes_{c\notin S_2} \mu^{(k+c)} .
    \end{equation}
\end{lemma}
\noindent
An important implication is the following pair of second moment bounds for Hermite polynomials under Gaussians with non-zero mean:
\begin{lemma}\label{lem:lower-bound-test}
For any $\mu \in \mathbb{R}$ and integer $k\ge \mu^2$, we have
    \[\Ex_{X \sim N(0,1)} h_k^2(X + \mu) = k!\cdot \exp(\Theta(\sqrt{k} |\mu|))\]
\end{lemma}
\begin{proof}
    By explicit calculation, we have:
    \begin{align*}
        \frac{1}{k!}\cdot \Ex_{X \sim N(0,1)} h_k^2(X+\mu) &= \sum_{j=0}^k \left(\binom{k}{j} \mu^{k-j} \right)^2 \cdot \frac{j!}{k!} \\
        &= \sum_{j=0}^k \frac{k!}{((k-j)!)^2\cdot j!} \mu^{2(k-j)} \\
        &= \sum_{j=0}^k \frac{k!}{(j!)^2\cdot (k-j)!} \mu^{2j} \\
        &= \sum_{j=0}^k \binom{k}{j}\frac{1}{j!}\cdot \mu^{2j}.
    \end{align*}

    To show the upper bound, since $\binom{k}{j} \le \frac{k^j}{j!}$, we have that
    \[
        \frac{1}{k!}\cdot \Ex_{X\sim N(0,1)} h_k^2(X + \mu) \le \sum_{j=0}^k \frac{(k\mu^2)^j}{(j!)^2} \le I_0(2\sqrt{k} |\mu|).
    \]
    where $I_0(x) = \sum_{j=0}^{\infty} \frac{(\frac{1}{4}x^2)^j}{(j!)^2}$ is the modified Bessel function of the first kind. 
    Then, since $I_0(x) = (1 + o(1)) \frac{e^x}{\sqrt{2\pi x}}$ (see e.g. 10.30 of \cite{olver2010nist}), this implies the upper bound.

    To obtain the lower bound, we let $j = \lceil \frac{\sqrt{k}|\mu|}{2} \rceil (\le k/2)$. 
    Then, we have that
    \[
        \binom{k}{j}\frac{1}{j!} \mu^{2j} \ge \frac{(k-j)^j\cdot \mu^{2j}}{j^{2j}} \ge \left(\frac{k\mu^2}{2j^2}\right)^j \ge 2^j = \exp(\Omega(\sqrt{k}|\mu|)),
    \] 
    which gives us the lower bound.

\end{proof}

\begin{lemma}\label{lem:hermite-tensor-upper}
    \[
        \left\|\Ex_{X\sim N(\mu,I)}[H_k(X)\otimes H_k(X)]\right\|_2 = k! \exp(O(\sqrt{k}\left\|\mu\right\|)).
    \]
\end{lemma}
\begin{proof}
By \cref{lem:hermite-second-moment},
\[
    \Ex_{X \sim N(\mu,I)}[H_k(X) \otimes H_k(X)] = \sum_{S_1,S_2\subset [k],|S_1|=|S_2|}\sum_{\substack{\text{Matchings } P \\ \text{of } S_1, S_2}} \bigotimes_{\{a,b\}\in P} I^{(a,k+b)} \bigotimes_{c\notin S_1} \mu^{(c)} \bigotimes_{c\notin S_2} \mu^{(k+c)} .
\]

Taking the operator norm on both sides, we get
\begin{align*}
    \left\|\Ex_{X\sim N(\mu,I)}[H_k(X)\otimes H_k(X)]\right\| &\le \sum_{j=0}^k \sum_{S_1,S_2\subset [k],|S_1|=|S_2|=j}\sum_{\substack{\text{Matchings } P \\ \text{of } S_1, S_2}} \left\|\mu\right\|^{2(k-j)} \\
    & = \sum_{j=0}^k \binom{k}{j}^2 j! \cdot \left\|\mu\right\|^{2(k-j)} \\
\end{align*}
which is same to the term in \cref{lem:lower-bound-test} up to factor $k!$, so we get this is upper bounded by $k!\cdot \exp(\Theta(\sqrt{k}\left\|\mu\right\|))$ as desired.
\end{proof}

%% file: exhaustive.tex
Suppose we receive an $\alpha$-pure set of $n$ samples where the inliers are distributed according to $N(\mu,I)$, and we know a bound $\| \mu \| \le r$. In this section, we will focus on how to construct a set $\mu_1,\dots,\mu_l$ such that some $\mu_i$ is $\eps$-close to $\mu$, all $\mu_i$ have a subset of $\alpha n$ points where the centered moments are close to $N(0,I)$, and all pairs $\mu_i,\mu_j$ are $\eps/2$-separated. 

Our ingredients from other sections will later imply that such a list has bounded size. Further, when we invoke the following theorem, we will be invoking it for a learned subspace of small dimension, so the dependence on $d$ is not as concerning as it may appear at first glance.

\begin{theorem}\label{thm:exhaust}
    Suppose there are $n$ samples $X_1,\dots,X_n \in \mathbb{R}^d$, $\alpha n$ of which are sampled according to $X_i \sim N(\mu,I)$. Consider parameters $0 < \alpha, \delta, \Delta < 1$, \, $\| \mu \| \le r$,\, $0 < \eps \le r$, and a positive integer $k^*$. There exists an algorithm that outputs a list $\mu_1,\dots,\mu_l$ such that the $\mu_i$ are $\eps/2$-separated, each $\mu_i$ has a corresponding vector $w^{(i)}_1,\dots,w^{(i)}_n \in [0,1]$ such that $\sum_{j=1}^n w^{(i)}_j = \alpha n$, and
    \begin{equation*}
        \left\| \frac{1}{\alpha n}\sum_{i=1}^{n} w_j^{(i)} \cdot (X_j - \mu_i)^{\otimes k} - \Ex_{Z \sim N(0,I)}[Z^{\otimes k}]\right\|_F \le \Delta \quad\forall k \in \{1,\dots,k^*\}.
    \end{equation*}
    Moreover, with probability at least $1-\delta$, there exists a $\mu^* \in \mu_1,\dots,\mu_l$ such that $\|\mu - \mu^* \|_2 \le \eps$. The algorithm has running time $O(\frac{r}{\eps \Delta})^d \cdot (2dk^*)^{Cdk^*} \cdot (\frac{1}{\alpha \delta \Delta})^C$ and uses $n =  \lceil\frac{4 (k^*)^{1.5} (8dk^*)^{k^*}}{\alpha \delta \Delta }\rceil $ samples.
\end{theorem}
\begin{proof}
\textbf{Overview.} Our plan is to consider an $\eps'$-cover $\mathcal{N}$ of the ball $\mathcal{B}$ in which $\mu$ is restricted, where $\eps' << \eps/2$. For each center in the cover, we will check whether there exists a subset of samples where the moments around the center look sufficiently close to Gaussian (in terms of the Frobenius norm; this may be computed efficiently, we discuss later). For all possible centers that pass this test, we will iterate in an arbitrary order, and greedily add a center to our set if it is not $\eps/2$-close to any center already in our set. Clearly, our final set will consist of centers where their moments satisfy our condition, and the centers are $\eps/2$-separated. What remains is to show that one of these centers will be $\eps$-close to the true $\mu$. Let $\mu^*$ be an arbitrary point in $\mathcal{N}$ that is $\eps'$-close to $\mu$. We will show that with probability at least $1-\delta$, $\mu^*$ will satisfy the moment conditions. Hence, either $\mu^*$ will be in our set, or $\mu^*$ is $\eps/2$-close to an item in our set. Since $\mu^*$ itself is $\eps'$-close to $\mu$ (where $\eps' \le \eps/2$), this immediately implies $\mu$ is $\eps$-close to an item in our set.

\textbf{$\eps'$-cover construction. } Let us define a simple $\eps'$-cover $\mathcal{N}$ for $\mathcal{B}$. Since each coordinate is within a bounded range of width $2r$, for each coordinate let us consider possible values $\{-r,\dots,-2\eps'/\sqrt{d},-\eps'/\sqrt{d},0,\eps'/\sqrt{d},2\eps'/\sqrt{d},\dots,r\}$; there are at most $1 + 2 \lceil r \sqrt{d}/\eps' \rceil \le 5 r \sqrt{d}/\eps'$ options. We will choose our net $\mathcal{N}$ to be the set of all $\le (5r \sqrt{d}/\eps')^d$ options that are within the ball. This is a valid $\eps'$ net, because for any $p \in \mathcal{B}$, consider the point $p'$ where each coordinate is rounded to the nearest value in the net that is closer to $0$: we observe $p' \in \mathcal{B}$ and $\|p - p' \|_2 \le \eps'$. Enumerating over this net takes $O(r\sqrt{d}/\eps')^d$ time.

\textbf{Moment-checking procedure. } For any fixed $\mu' \in \mathcal{N}$, we hope to find a collection of values $w_1,\dots,w_n \in [0,1]$ such that $\sum_i w_i = \alpha n$, and the following condition holds for the moment tensors:
\begin{equation*}
    \left\| \frac{1}{\alpha n}\sum_{i=1}^{n} w_i \cdot (X_i - \mu')^{\otimes k} - \Ex_{Z \sim N(0,I)}[Z^{\otimes k}]\right\|_F \le \Delta \quad\forall k \in \{1,\dots,k^*\}
\end{equation*}

This condition would be implied by
\begin{equation}
    \sum_{k=1}^{k^*} \left\| \frac{1}{\alpha n}\sum_{i=1}^{n} w_i \cdot (X_i - \mu')^{\otimes k} - \Ex_{Z \sim N(0,I)}[Z^{\otimes k}]\right\|_F^2 \le \Delta^2. \label{eq:condition-check}
\end{equation}

This condition is quite convenient to check: the constraints on $w$ are linear constraints, and the objective is simply a least-squares regression task, so we may run in time $\poly(n,d^{k^*})$ for a fixed $\mu'$.

\textbf{Showing $\mu^*$ passes the moment-checking procedure. } Recall $\mu^*$ is an arbitrary point in our set $\mathcal{N}$ that is $\eps'$-close to $\mu$. We desire to show that with probability at least $1-\delta$, $\mu^*$ will pass the moment-checking procedure. Our plan is to show that choosing the $w_i$ corresponding to the $\alpha n$ inliers sampled from $N(\mu,I)$ will pass the test. Our moment-checking condition will certainly pass if for all $k \in \{1,\dots,k^*\}$, it holds:
\begin{equation}
    \left\| \frac{1}{\alpha n}\sum_{i \in \textrm{$\alpha n$ inliers}} (X_i - \mu^*)^{\otimes k} - \Ex_{Z \sim N(0,I)}[Z^{\otimes k}]\right\|_F \le \Delta/\sqrt{k^*}
\end{equation}
where $X_i \sim N(\mu,I)$. This is equivalent to the following condition for $Y_1,\dots,Y_{\alpha n} \sim N(\mu-\mu^*,I)$:
\begin{equation}
    \left\| \frac{1}{\alpha n}\sum_{i =1}^{\alpha n} Y_i^{\otimes k} - \Ex_{Z \sim N(0,I)}[Z^{\otimes k}]\right\|_F \le \Delta/\sqrt{k^*}\label{eq:small-mu-cond}
\end{equation}
The following technical lemma allows us to show this condition (\cref{eq:small-mu-cond}) occurs with high probability when $\|\mu - \mu^*\|$ is small enough (proof deferred to \cref{app:frob-diff}).
\begin{lemma}\label{lemma:frob-diff}
    For parameters $0 < \delta,\gamma < 1$, suppose it holds that $m \ge \frac{4 (8dk)^k}{\delta \gamma}$ and $\| \mu \|_\infty \le \frac{\gamma}{2 (4dk)^{k/2}}$. Then,
    \begin{equation*}
        \Pr_{X_1,\dots,X_m \sim N(\mu,I)}\left[ \left\| \frac{1}{m} \sum_{i=1}^m X_i^{\otimes k}  - \Ex_{Z \sim N(0,I)}[Z^{\otimes k}] \right\|_F  \ge \gamma \right] \le \delta.
    \end{equation*}
\end{lemma}

This immediately lets us conclude that \cref{eq:small-mu-cond} holds with probability at least $1-\delta$:
\begin{corollary}
    The moment-checking condition will hold for $\mu^*$ with probability at least $1-\delta$ if $n \ge \frac{4 (k^*)^{1.5} (8dk^*)^{k^*}}{\alpha \delta \Delta }$ and $\eps' \le \frac{\Delta}{ 2\sqrt{k^*}(4dk^*)^{k^*/2}}$.
\end{corollary}
\begin{proof}
    We invoke \cref{lemma:frob-diff} for each $k \in \{1,\dots,k^*\}$ with $\gamma = \Delta/\sqrt{k^*}$, $\delta' = \delta/k^*$, $\|\mu \|_{\infty} \le \eps'$, and $m = \alpha n$. The first condition holds since 
    \begin{equation*}
        m \ge \frac{4 (8dk)^k}{\delta' \gamma } \impliedby n \ge \frac{4 (k^*)^{1.5} (8dk^*)^{k^*}}{\alpha \delta \Delta }.
    \end{equation*}
    The second condition holds since
    \begin{equation*}
        \| \mu \|_\infty \le \frac{\gamma}{2 (4dk)^{k/2}} \impliedby \eps' \le \frac{\Delta}{ 2\sqrt{k^*}(4dk^*)^{k^*/2}}.
    \end{equation*}
\end{proof}

\textbf{Concluding. } Hence, our algorithm succeeds with probability at least $1-\delta$ with our choice of parameters $\eps' = \min(\eps/2, \frac{\Delta}{ 2\sqrt{k^*}(4dk^*)^{k^*/2}})$ and $n = \lceil\frac{4 (k^*)^{1.5} (8dk^*)^{k^*}}{\alpha \delta \Delta }\rceil $. Our running time is $O(r\sqrt{d}/\eps')^d \cdot \poly(n,d^{k^*}) = O(\frac{r}{\eps \Delta})^d \cdot (2dk^*)^{Cdk^*} \cdot (\frac{1}{\alpha \delta \Delta})^C$. This concludes the proof of our theorem.
\end{proof}

%% file: fool-moments.tex
The guarantees from Theorem 5.6 of \cite{gollakota2023moment} are lossy in the setting where moments are sub-gaussian and the number of halfspaces are super-constant, as is the case in our application. We provide a sharper result (proof deferred to \cref{app:foolproof}):

\begin{lemma}\label{lemma:fool}
    Suppose two distributions $G,T$ over $\mathbb{R}^d$ match moments up to order $k$ (which is even), meaning
    \begin{equation*}
        \left| \Ex_G[\langle G, v \rangle^i] - \Ex_T[\langle T,v \rangle^i] \right| \le \Delta
    \end{equation*}
    for all $\|v\|_2=1$ and $i \in \{1,\dots, k \}$. Moreover, suppose $G$ has subgaussian moments where
    \begin{equation*}
        \Ex_{G}[\langle G,v\rangle^i] \le (C_1 \sqrt{i})^i
    \end{equation*}
    and its linear projections are anticoncentrated such that
    \begin{equation*}
        \Ex_G[\langle G,v \rangle \in [L,R]] \le C_2 \cdot (R-L). 
    \end{equation*}
    for constants $C_1,C_2>0$. Consider any collection of halfspaces $H_1,\dots,H_m$ where $H_i \triangleq \mathbbm{1}[\langle x, a_i \rangle \ge b_i]$. There exist constants $C_3,C_4 \ge 1$ where if $k \ge C_3 \cdot \max(m^6/\eps^2,\log^8(1/\eps)/\eps^2)$, $\Delta \le \left( \frac{1}{C_4 k} \right)^k$, and $0 < \eps \le \nicefrac{1}{2}$, then for the intersection $H \triangleq \bigcap_{i=1}^m H_i$ it holds
    \begin{equation*}
        \left| \Pr_G[X \in H] - \Pr_T[X \in H] \right| \le C   \eps .
    \end{equation*}
\end{lemma}

%% file: combining.tex
Let us recall our general plan. First, we will use prior work of \cite{diakonikolas2022list} to reduce the task to $O(1/\alpha)$ subproblems where $\|\mu\|_2 \le O(\sqrt{\log(1/\alpha)})$. Next, for each subproblem, we will learn the low-dimensional subspace via \cref{subsec:subspace}. Afterwards, we will exhaustively search over a net within this subspace via \cref{subsec:exhaust}, where this subroutine returns $\mu_i$ satisfying the moment matching condition, and one of them is close to $\mu$, yet we have not bounded the size of this output list. Finally, we will combine our moment-matching fooling result in \cref{subsec:fool} with the ideas in our information-theoretic proof of \cref{thm:info-theory} to prove the list must have bounded size. We now combine all the ingredients.

Throughout this algorithm, we will need three groups of samples of size $n_1,n_2,n_3$ such that $n = n_1+n_2 + n_3$.  We will randomly distribute our $n$ samples into these three groups, and since $n$ is sufficiently large, the number of good inliers in each group will be at least $\alpha n_i /2$ with high probability. For the remaining description of our algorithm subroutines, we will leverage $\alpha/2$ as our inlier parameter.

\textbf{Using prior work of Diakonikolas, Kane, Karmalkar, Pensia, and Pittas \cite{diakonikolas2022list}.} For any constant $\delta > 0$, Theorem 1.2 of \cite{diakonikolas2022list} outputs a list of size $O(1/\alpha)$, that with probability at least $1-\delta$ contains a point within distance $C \sqrt{\log(1/\alpha)}$ of the true $\mu$. Moreover, this algorithm runs in time $(d \log(1/\alpha))^{C \log(1/\alpha)}$ and uses $n_1 = (d \log(1/\alpha))^{C \log(1/\alpha)}$ samples. From this output list, we will focus on $O(1/\alpha)$ subproblems, where after translation we may assume that $\|\mu\|_2 \le C \sqrt{\log(1/\alpha)}$. We now focus on each of the subproblems separately.

\textbf{Learning a low-dimensional subspace.} We invoke \cref{thm:subspace-learning} with $\alpha' = \alpha/2$ and $\eps' = \eps/2$. This gives a subspace with dimension at most $\exp(C \log^2(1/\alpha) /\eps^2 \cdot(\log(\log(1/\alpha)/\eps)))$, where $\mu$ is within $\eps/2$ of the subspace with constant probability. We may run this constantly many times on different samples to boost the failure probability to any small constant, while keeping the subspace dimension at most $\exp(C \log^2(1/\alpha) /\eps^2 \cdot(\log(\log(1/\alpha)/\eps)))$. Learning the subspace has sample/time complexity $(d \log(1/\alpha)/\eps)^{C \log^2(1/\alpha)/\eps^2}$.

\textbf{Exhaustively searching within the low-dimensional subspace. } We now invoke \cref{thm:exhaust} to search the low-dimensional subspace. We may use $d' = (C \log^2(1/\alpha)/\eps^2)^{C \log^2(1/\alpha)/\eps^2}$, $\eps = \eps'/2$, $r \le C \sqrt{\log(1/\alpha)}$, and $\delta$ to be a sufficiently small constant. We defer the choice of $\Delta$ and $k^*$ until later. All $n_3$ samples will be input to the algorithm after projecting to the subspace. Observe how with probability at least $1-\delta$, the list contains a $\mu^*$ in the subspace where $\|\mu^* - \Pi_{V}\mu\|_2 \le \eps/2$, and hence $\|\mu - \mu^* \| \le \eps$. Moreover, all $\mu_i$ are $\eps/4$-separated. The main remaining question is to bound the size of this list. 

\textbf{Bounding the list size. } Let $\mu_1,\dots,\mu_l$ denote the output list from some call to the exhaustive search subroutine. We will show how the list size must be bounded by using similar ideas to our information-theoretic proof \cref{thm:info-theory}, in addition to our moment-matching fooling technical result \cref{lemma:fool}. Let $r_0 = C \sqrt{\log(1/\alpha)}$ be the bound on the error from the algorithm of \cite{diakonikolas2022list}. For non-negative integer $i$, let $|M_i|$ denote the minimum size $r_i\triangleq \frac{r_0}{2^{i}}$-cover of $\mu_1,\dots,\mu_l$. If we choose $\Delta$ and $k^*$ appropriately, we may bound the ratio of consecutive $|M_i|/|M_{i-1}|$. Eventually, for some $i^*$ where $r_{i^*} < \eps/8$, we may conclude $|M_{i^*}| = l$ because all $\mu_i$ are $\eps/4$-separated. Since the ratio bound will let us bound $|M_{i^*}|$, this will in turn bound $l$. We now prove the ratio bound:
\begin{claim}
    For any positive integer $i$ where $r_i \ge \frac{\eps}{100}$, it holds that $\frac{|M_i|}{|M_{i-1}|} \le \exp(C \log^2(1/\alpha) / r_i^2)$.
\end{claim}
\begin{proof}
    This proof mirrors the proof of \cref{claim:m-bound}, except we must adjust that our $\mu_i$ only satisfy moment-matching conditions, instead of the $\alpha$-consistent property. 

    We will construct a cover $M_i$ by considering covering each $r_{i-1}$-radius ball of $M_{i-1}$ separately. The size of the cover for each ball is bounded by the maximum-possible size subset $\mu_1,\dots,\mu_{l'}$ such that all $\mu_i$ are $r_{i}$-separated. We will prove that such a set cannot exist for $l'$ too large. After translating all such $\mu_i$ accordingly, recall all $\|\mu_i\|_2 \le r_{i-1}$. Let us again define corresponding regions $R_i$ where the density of $\mu_i$ is maximal: $R_i \triangleq \{ \|x - \mu_i \| < \|x - \mu_j \| \}$. Let $D$ denote the normalized empirical distribution over the $n_3$ samples. Similarly, let $D_i$ denote the normalized empirical distribution corresponding to each $w^{(i)}$. Since the total mass of $D$ is 1, we know:
    \begin{equation}\label{eq:finite-r}
        \sum_{i=1}^{l'}\Pr_{X \sim D}[X \in R_i] \le 1 \implies \sum_{i=1}^{l'} \alpha' \Pr_{X \sim D_i}[X \in R_i] \le 1 \implies \sum_{i=1}^{l'} \frac{\alpha}{2} \Pr_{X \sim D_i}[X \in R_i] \le 1
    \end{equation}
    We now want to show that $\Pr_{X \sim D_i}[X \in R_i] \approx \Pr_{X \sim N(\mu_i,I)}[X \in R_i]$. Note how we may translate the distribution $D_i$ by $-\mu_i$ (and similarly define $R'_i$ as the region $R_i$ with the same translation), so we instead study
    \begin{equation*}
        |\Pr_{X \sim D_i}[X \in R_i] - \Pr_{X \sim N(\mu_i,I)}[X \in R_i]| = |\Pr_{X \sim (D_i - \mu_i)}[X \in R'_i] - \Pr_{N(0,I)}[R_i']|
    \end{equation*}
    This is a quantity that we may bound via our moment-matching fooling technical result of \cref{lemma:fool}. Note how we are aiming to show contradiction for a sufficiently large $l'$; we will choose $l'$ so that is always at most $\exp(C \log^2(1/\alpha) / r_i^2)$. We now want to use \cref{lemma:fool} such that each $D_i$ will fool an intersection of at most $l' \le \exp(C \log^2(1/\alpha) / r_i^2)$ halfspaces with additive error at most $\frac{1}{l'}$. In the language of \cref{lemma:fool}, $G = N(0,I)$, and $H$ is the distribution $D_i$ translated by $-\mu_i$. The subgaussian moments assumption and anticoncentration assumption follows immediately since $G= N(0,I)$. The lemma implies fooling halfspaces up to error $\frac{1}{l'}$ if we choose $k^* = \exp(C \log^2(1/\alpha) / \eps^2) \ge \exp(C \log^2(1/\alpha) / r_i^2)$ and $\Delta = \exp(C \log^2(1/\alpha) / \eps^2)^{-\exp(C \log^2(1/\alpha) / \eps^2)}$. Using this, we continue:
    \begin{align*}
        \cref{eq:finite-r} &\iff  \sum_{i=1}^{l'} \frac{\alpha}{2} \left(\Pr_{X \sim N(\mu_i,I)}[X \in R_i] + \left(\Pr_{X \sim D_i}[X \in R_i] - \Pr_{X \sim N(\mu_i,I)}[X \in R_i]\right)\right)\le 1 \\
        & \implies \sum_{i=1}^{l'} \frac{\alpha}{2} \Pr_{X \sim N(\mu_i,I)}[X \in R_i] \le 1 + \frac{\alpha}{2} \\
        & \implies  \sum_{i=1}^{l'} \Pr_{X \sim N(\mu_i,I)}[X \in R_i]  \le \frac{4}{\alpha} \\
        & \iff  \sum_{i=1}^{l'} q_i  \le \frac{4}{\alpha} \tag{recalling the definition of $q_i$} \\
        & \implies \poly(\alpha) \cdot 2^{\frac{c r_{i}^3}{r_{i-1}^2} \sqrt{\log l'}} \le \frac{4}{\alpha} \tag{invoking \cref{thm:q-bound} with $r = r_{i-1}$ and $\beta = r_i$}\\
        & \iff \poly(\alpha) \cdot 2^{c r_i \sqrt{\log l'}} \le \frac{4}{\alpha} \\
        & \implies l' \le \exp(C \log^2(1/\alpha) / r_i^2).
    \end{align*}
    This completes the proof of our claim, since the maximal size $r_i$-separated set inside each $r_{i-1}$-radius ball must have size at most $\exp(C \log^2(1/\alpha) / r_i^2)$.
\end{proof}

With this ratio bound in hand, the observation that $l = |M_{i^*}|$, and setting $i^* = \lceil \log_2(r_0 / (\eps/16)) \rceil$, we conclude
\begin{align*}
    t &= |M_{i^*}| \le |M_0| \cdot \Pi_{i=1}^{i^*} \frac{|M_i|}{|M_{i-1}|} = \Pi_{i=1}^{i^*} \frac{|M_i|}{|M_{i-1}|} \\
    & \le \Pi_{i=1}^{i^*} \exp(C \log^2(1/\alpha) / r_i^2) \le \exp(C \log^2(1/\alpha) / r_{i^*}^2) \\
    & \le \exp(C \log^2(1/\alpha) / \eps^2).
\end{align*}

This bounds the size of each of the list returned in each of the $C/\alpha $ subproblems, and in total we attain a list size bound of $(C/\alpha) \cdot \exp(C \log^2(1/\alpha) / \eps^2) \le \exp(C \log^2(1/\alpha) / \eps^2)$.

\textbf{Concluding runtime and sample usage. } Finally, we revisit runtime and sample complexity now that all parameters have been chosen (namely, $\Delta$ and $k^*$). The algorithm of \cite{diakonikolas2022list} still has sample/time complexity of $(d \log(1/\alpha))^{C \log(1/\alpha)}$. Learning the subspace still has sample/time complexity $(d \log(1/\alpha)/\eps)^{C \log^2(1/\alpha)/\eps^2}$. The exhaustive search has sample complexity $\exp(\exp(C\log^2(1/\alpha)/\eps^2))$ and time complexity $\exp(\exp(C \log^2(1/\alpha) /\eps^2 \cdot(\log(\log(1/\alpha)/\eps))))$. In total, the sample complexity is $d^{C \log^2 (1/\alpha)/\eps^2} + \exp(\exp(C\log^2(1/\alpha)/\eps^2))$, and the time complexity is $d^{C \log^2 (1/\alpha)/\eps^2} + \exp(\exp(C \log^2(1/\alpha) /\eps^2 \cdot(\log(\log(1/\alpha)/\eps))))$.

%% file: technical-lemmas.tex
\section{Deferred proofs}

\subsection{Proof of \cref{cor:semiverified}}\label{app:semiverified}
First, we run our list-decodable mean estimation algorithm from \cref{thm:efficient} and obtain a list $\mu_1,\dots,\mu_l$ where $l \le L$. Let $\mu^*$ be any arbitrary item in the list where $\| \mu - \mu^* \|_2 \le \eps$.  Given this list, the rest of the proof follows from standard ideas.

We now design a subroutine where, given two items $\mu_i,\mu_j$ such that $\| \mu_i - \mu_j \|_2 \ge  4 \eps$, we will output a ``winner'' between $\mu_i$ and $\mu_j$. Our desired guarantee is that for all such pairwise tests, if one of $\mu_i,\mu_j$ is $\mu^*$, then $\mu^*$ will always win.

Our test is simple. For each pairwise comparison, we use our $n_2$ samples $X_1,\dots,X_{n_2}$, and transform them into $Y_k = \langle X_k - \frac{\mu_i + \mu_j}{2}, \frac{\mu_i - \mu_j}{\| \mu_i - \mu_j \|_2} \rangle$. Then, our test will output $\mu_i$ if the average $\frac{1}{n_2} \sum_{k=1}^{n_2} Y_k > 0$, and $\mu_j$ otherwise. 

We will show that $\mu^*$ wins all its tests with high probability. Without loss of generality, suppose $\mu^*$ is $\mu_i$, we prove that $\frac{1}{n_2}\sum_{k=1}^{n_2} Y_k > 0$ with probability at least $\delta/|L|$. Observe how the $Y_k$ are distributed according to $N(\langle \mu - \frac{\mu^* + \mu_j}{2}, \frac{\mu^* - \mu_j}{\|\mu^* - \mu_j\|_2} \rangle,1)$. The mean of this univariate Gaussian distribution is at least
\begin{align*}
    \langle \mu - \frac{\mu^* + \mu_j}{2}, \frac{\mu^* - \mu_j}{\|\mu^* - \mu_j\|_2} \rangle &= \langle \mu^* + (\mu - \mu^*) - \frac{\mu^* + \mu_j}{2}, \frac{\mu^* - \mu_j}{\|\mu^* - \mu_j\|_2}\rangle \\
    & \ge \langle \frac{\mu^* - \mu_j}{2}, \frac{\mu^* - \mu_j}{\|\mu^* - \mu_j\|_2}\rangle - \eps \\
    & = \frac{\|\mu^* - \mu_j \|_2}{2} - \eps \ge \eps
\end{align*}

With probability at least $1-\delta/|L|$, the empirical mean will concentrate within $\eps/2$ of its true mean from $n_2 \ge \frac{C \log(|L|/\delta)}{\eps^2}$ samples. By union bound, $\mu^*$ will pass all its tests with probability at least $1-\delta$. 

Finally, we output any $\mu_i$ that passes all its tests. In the event where $\mu^*$ was undefeated, this means $\mu_i$ is within $4\eps$ of $\mu^*$, and hence within $8\eps$ of $\mu$. If we rescale $\eps$ by a factor of $8$, then we obtain the desired guarantee. 

\subsection{Proof of \cref{lemma:isoperimetry}}\label{app:isoper}
\begin{proof}
Denote $\mu(A)$ by $p(\le 1/2)$. We define $x(\ge 0)$ be the solution of $\int_x^{\infty} \frac{1}{\sqrt{2\pi}} e^{-t^2/2}\:dt = p$.

Gaussian isoperimetry inequality tells us,
\[
    \mu(A_\epsilon) \ge \int_{x-\epsilon}^{\infty} \frac{1}{\sqrt{2\pi}} e^{-t^2/2}\:dt = \int_{x}^{\infty} \frac{1}{\sqrt{2\pi}} e^{-(t-\epsilon)^2/2}\:dt
\]

We know that when $p=1/3$, $x \approx 0.43$, so when $p\ge 1/3$, we have $x\le 0.44$, and when $p\le 1/3$, we have $x\ge 0.42$ and by Mills ratio, we know $p\ge \frac{x}{1+x^2}\frac{e^{-x^2/2}}{\sqrt{2\pi}} \ge \frac{e^{-x^2/2}}{10x}$, so $x\ge \Omega(\sqrt{\log(1/p)})$. Therefore in either case we have $x(p) = \Omega(\sqrt{\log(1/p)}) - O(1)$, so
\[
    \frac{\mu(A_\epsilon)}{\mu(A)} \ge \min_{t\ge x}{\frac{e^{-(t-\epsilon)^2/2}}{e^{-t^2/2}}} = \frac{e^{-(x-\epsilon)^2/2}}{e^{-x^2/2}} = e^{-\epsilon^2/2} \cdot e^{x\epsilon} = \exp(-\epsilon^2 /2- O(\epsilon) + \Omega(\epsilon \sqrt{\log(1/p)})),
\]
which right hand side is $\Omega(1)\exp(\Omega(\epsilon \sqrt{\log(1/p)}))$ when $\eps = O(1)$ and is $\exp(-O(\epsilon^2) + \Omega(\epsilon \sqrt{\log(1/p)}))$ when $\eps = \Omega(1)$, so the lemma follows.
\end{proof}

\subsection{Proof of \cref{lemma:lb}}\label{app:lb}
\begin{proof}
    Let $D_0$ be the distribution $\frac{1}{2} N(0,I)$, and let $e_i$ be the $d$-dimensional vector that has a $1$ in the $i$-th coordinate and $0$ for all other coordinates. Additionally, for $i \in \{1,\dots,d\}$, let $D_i$ be the function $D_i(x) = \max(0,\alpha [N(2\eps e_i, I)](x) - D_0(x))$. 
    
    Suppose the total mass of each $D_i$ is at most $\frac{1}{2 d}$. In this case, we could define our distribution $D$ such that its density is at least $D(x) \ge \sum_{i=0}^d D_i(x)$ everywhere, and any valid list must have size at least $d$ since all $2\eps e_i$ are $\alpha$-consistent and $>2\eps$-separated. Our plan is to analyze the total mass for any $D_i$, and then choose $d$ accordingly.

    First, let us bound the region where $D_i$ is nonzero. This occurs when the ratio $\frac{D_0(x)}{\alpha[N(2 \eps e_i,I)](x)} < 1$. We analyze this ratio:
    \begin{align*}
        \frac{D_0(x)}{\alpha [N(2 \eps e_i,I)](x)} &= \frac{1}{2\alpha} e^{(- \|x\|_2^2 + \|x - 2\eps e_i \|_2^2)/2} \\ 
        &= \frac{1}{2 \alpha} e^{- \langle x, 2\eps e_i\rangle + \| 2 \eps e_i \|_2^2/2}\\
        &= \frac{1}{2 \alpha} e^{- 2 \eps x_i + \eps^2 /2 }
    \end{align*}
    Hence, we may observe conditions under which $D_i$ is nonzero:
    \begin{equation*}
        D_i(x) > 0\iff x_i > \frac{\eps^2/2 + \log(\nicefrac{1}{2\alpha})}{\eps}
    \end{equation*}
    From here, we may bound the total mass of $D_i$:
    \begin{align*}
        \int_{\mathbb{R}^d} D_i(x) &\le \Pr_{X \sim N(2 \eps e_i, I)}\left[x_i > \frac{\eps^2/2 + \log(\nicefrac{1}{2\alpha})}{\eps}\right]\\
        &= \Pr_{X \sim N(0,1)}\left[x > \frac{\eps^2/2 + \log(\nicefrac{1}{2\alpha})}{\eps} - 2\eps\right]\\
        & \le \Pr_{X \sim N(0,1)}\left[x > \frac{\log(\nicefrac{1}{2\alpha})}{2\eps}\right] \tag{using $\eps < \frac{\sqrt{\log(\nicefrac{1}{2\alpha})}}{2}$}\\
        & \le \frac{2 \eps}{\log(\nicefrac{1}{2\alpha})} \cdot e^{-\frac{\log^2(\nicefrac{1}{2\alpha})}{8\eps^2}} \le e^{-\frac{\log^2(\nicefrac{1}{2\alpha})}{8\eps^2}}
    \end{align*}
    Hence, we may choose any $d$ (and thus force any list size) where
    \begin{equation*}
        d \le \max(1, \lfloor e^{\frac{\log^2(\nicefrac{1}{2\alpha})}{8\eps^2}} / 2 \rfloor ).
    \end{equation*}
    This means we can force a list of size $\exp(c \log^2(\nicefrac{1}{2\alpha}) / \eps^2)$.
\end{proof}

\subsection{Omitted proofs from Section~\ref{sec:subspace}}

\begin{lemma}\label{lem:hermite-concentration}
    \[\Pr_{X_1,\dots,X_m \sim N(\mu,1)}\left[ \left\| \frac{1}{m} \sum_{i=1}^m H_k(X_i) \otimes H_k(X_i)  - \Ex_{X \sim N(\mu,I)}[H_k(X_i) \otimes H_k(X_i)] \right\|_F \ge a \right] \le \frac{(dk\max(1,\|\mu \|_{\infty}))^{O(k)}}{a^2m},\] 
\end{lemma}
\begin{proof}[Proof of Lemma~\ref{lem:hermite-concentration}]
    By
    \begin{equation}\label{eqn:hermite-tensor-expand}
        H_k(X) = \sum_{\substack{P \text{ Partition of set } S \\ \text{subset with size 1 and 2}}} \bigotimes_{\{a,b\}\in P} (-I)^{(a,b)} \bigotimes_{\{c\}\in P} X^{(c)}.
    \end{equation}

    we know $H_k(X) \otimes H_k(X)$ can written as the sum of terms in the form of \[\bigotimes_{\{a,b\}\in P} (-I)^{(a,b)} \bigotimes_{\{c\}\in P} X^{(c)} \bigotimes_{\{a,b\}\in P'} (-I)^{(a+k,b+k)} \bigotimes_{\{c\}\in P} X^{(c+k)}.\]

    For each such term, using concentrated \cref{claim:frob-conc} and let $\theta = \max(1,\|\mu \|_{\infty})$, we know the difference between empirical average and the expectation is at most $a/(k!)^2$ with probability at least $1-\frac{(k!)^4(8dk \theta^2)^k}{a^2m} = 1 - \frac{(dk\theta)^{O(k)}}{a^2m}$.
    
    Since the number of summands is bounded by $(k!)^2$, by union bound we complete the proof.
    
\end{proof}

\begin{lemma}\label{lem:gaussian-concentration}
    For $\mu\in \mathbb{R}$, and integer $k\ge 0$, we have
    \[\Pr_{X_1,\dots,X_m \sim N(\mu,1)}\left[ \left\| \frac{1}{m} \sum_{i=1}^m f(X_i)  - \Ex_{X \sim N(\mu,I)}[f(X)] \right\|_2 \ge a \right] \le \frac{\exp(O(k+\sqrt{k}|\mu|))}{a^2m},\] 
    for $f = h_k^2 $ and $h_k^4$.
\end{lemma}
\begin{proof}
    By Chebyshev's inequality, it suffices to show $\Ex_{x\sim N(\mu,1)}[f^2(x)] \le \exp(O(k+\sqrt{k}|\mu|))$. Indeed by hypercontractivity, we have
    \[
        \left(\Ex_{x\sim N(\mu,1)}[h_k^8(x)]\right)^{1/8} \le e^{O(k)}\left(\Ex_{x\sim N(\mu,1)}[h_k^4(x)]\right)^{1/4} \le e^{O(k)} \left(\Ex_{x\sim N(\mu,1)}[h_k^2(x)]\right)^{1/2}
    \]
    which is at most $\exp(O(\sqrt{k}|\mu|))$ due to \cref{lem:lower-bound-test}, so the lemma follows.
\end{proof}

\subsection{Proof of \cref{lemma:frob-diff}}\label{app:frob-diff}
\begin{proof}
    We may bound the desired quantity by decomposing into two components:
\begin{equation}
    \left\| \frac{1}{m} \sum_{i=1}^m X_i^{\otimes k}  - \Ex_{Z \sim N(0,I)}[Z^{\otimes k}] \right\|_F \le \left\| \Ex_{Z \sim N(0,I)}[Z^{\otimes k}]  - \Ex_{X \sim N(\mu,I)}[X^{\otimes k}] \right\|_F + \left\| \frac{1}{m} \sum_{i=1}^m X_i^{\otimes k}  - \Ex_{X \sim N(\mu,I)}[X^{\otimes k}] \right\|_F\label{eq:frob-decompose}
\end{equation}

We now bound the first term:
\begin{claim}\label{claim:expect-frob}
    $\| \Ex_{z \sim N(0,I)}[(\mu + z)^{\otimes k}] - \Ex_{z \sim N(0,I)}[z^{\otimes k}] \|_F^2 \le (4d k)^{k/2} \|\mu \|_{\infty} $
\end{claim}
\begin{proof}
\begin{align*}
    & \| \Ex_{z \sim N(0,I)}[(\mu + z)^{\otimes k}] - \Ex_{z \sim N(0,I)}[z^{\otimes k}] \|_F^2\\
    & = \sum_{i_1,\dots,i_k \in [d]} \left(\Ex_{z \sim N(0,I)}[\Pi_{j=1}^k (\mu_{i_j} + z_{i_j})]- \Ex_{z \sim N(0,I)}[\Pi_{j=1}^k z_{i_j}]\right)^2 \intertext{Consider each monomial separately. The monomials of only $z$ terms will cancel out. Otherwise, a monomial will have at least one $\mu$ term. Since the expected value of a monomial of $z$ with at most $k-1$ terms is at most $(k-2)!!$, then:}
    & \le d^k \cdot \left((2^k-1) \cdot \|\mu \|_{\infty} \cdot (k-2)!!\right)^2 \\
    & \le (4d k)^k \|\mu \|_{\infty}^2 \\
    & \implies \| \Ex_{z \sim N(0,I)}[(\mu + z)^{\otimes k}] - \Ex_{z \sim N(0,I)}[z^{\otimes k}] \|_F \le (4d k)^{k/2} \|\mu \|_{\infty} \quad\qedhere
\end{align*}
\end{proof}
We may also bound the second term of \cref{eq:frob-decompose}:
\begin{claim}\label{claim:frob-conc}
    $\Pr_{X_1,\dots,X_m \sim N(\mu,I)}\left[ \| \frac{1}{m} \sum_{i=1}^m X_i^{\otimes k}  - \Ex_{X \sim N(\mu,I)}[X^{\otimes k}] \|_F \ge a \right] \le \frac{(8dk \cdot \max(1,\|\mu \|_{\infty}^2))^k}{a^2 m}$
\end{claim}
\begin{proof}
Using Chebyshev's inequality we observe
\begin{equation*}
    \Pr_{X_1,\dots,X_m \sim N(\mu,I)}\left[ \| \frac{1}{m} \sum_{i=1}^m X_i^{\otimes k}  - \Ex_{X \sim N(\mu,I)}[X^{\otimes k}] \|_F \ge a \right] \le \frac{\operatorname{Var}_{X \sim N(\mu,I)}(\|X^{\otimes k} \|_F)}{a^2}.
\end{equation*}

Hence, we will bound
\begin{align*}
    & \operatorname{Var}_{X \sim N(\mu,I)}(\|X^{\otimes k} \|_F) \le \Ex_{X_1,\dots,X_m \sim N(\mu,I)}\left[ \sum_{i_1,\dots,i_k \in [d]} \left( \frac{1}{m} \sum_{j=1}^m \left( \Pi_{\ell=1}^k (X_j)_{i_\ell} - \Ex_{X \sim N(\mu,I)}[\Pi_{\ell=1}^k X_{i_\ell}] \right)\right)^2\right] \\
    & = \frac{1}{m} \sum_{i_1,\dots,i_k \in [d]} \Ex_{X \sim N(\mu,I)} \left[ \left( \Pi_{\ell=1}^k X_{i_\ell} - \Ex_{Y \sim N(\mu,I)}[\Pi_{\ell=1}^k Y_{i_\ell}]\right)^2\right] \\
    & \le \frac{1}{m} \sum_{i_1,\dots,i_k \in [d]} \Ex_{X \sim N(\mu,I)} \left[ \left( \Pi_{\ell=1}^k X_{i_\ell}\right)^2\right] \\
    & = \frac{1}{m} \sum_{i_1,\dots,i_k \in [d]} \Ex_{Z \sim N(0,I)} \left[ \left( \Pi_{\ell=1}^k (Z_{i_\ell} + \mu_{i_\ell})\right)^2\right] \\
    & \le \frac{1}{m} \sum_{i_1,\dots,i_k \in [d]} 2^{2k} \cdot (2k-1)!! \cdot \max(1,\|\mu \|_{\infty}^{2k}) \\
    & \le \frac{(8dk \cdot \max(1,\|\mu \|_{\infty}^2))^k}{m}.
\end{align*}

All together, this implies our claim that
\begin{equation*}
    \Pr_{X_1,\dots,X_m \sim N(\mu,I)}\left[ \| \frac{1}{m} \sum_{i=1}^m X_i^{\otimes k}  - \Ex_{X \sim N(\mu,I)}[X^{\otimes k}] \|_F \ge a \right] \le \frac{(8dk\cdot \max(1,\|\mu \|_{\infty}^2))^k}{a^2 m}. \qedhere
\end{equation*}
\end{proof}

Concluding the proof of our lemma, we will set our parameters so that both terms of \cref{eq:frob-decompose} are at most $\gamma/2$ with probability at least $1-\delta$.

Bounding the first term with \cref{claim:expect-frob}, it is sufficient to set $\|\mu\|_{\infty}$ such that:
\begin{equation*}
    (4dk)^{k/2} \|\mu\|_{\infty} \le \gamma/2 \impliedby \|\mu\|_\infty \le \frac{\gamma}{2 (4dk)^{k/2}}
\end{equation*}
Bounding the second term with \cref{claim:frob-conc} and $\| \mu \|_\infty \le 1$, it is sufficient to set $m$ such that:
\begin{equation*}
    \frac{(8dk)^k}{(\gamma/2)^2 m} \le \delta \impliedby m \ge \frac{4 (8dk)^k}{\delta \gamma}.
\end{equation*}
This completes the proof of our lemma.
\end{proof}

\subsection{Proof of \cref{lemma:fool}}\label{app:foolproof}
\begin{proof}
    The proof will mostly follow the structure of \cite{diakonikolas2010bounded}, while making adjustments that allow for moments matching \textit{approximately} instead of exactly in their work.

    Consider the transformation $F: \mathbb{R}^d \rightarrow \mathbb{R}^m$ where $F(x)_i = \langle x, a_i \rangle$. Then the region of intersection for halfspaces is $\mathbbm{1}[F(x)_i \ge b_i \, \, \forall i \in [m]]$. Without loss of generality, all $\|a_i\|_2=1$.

    Let $I_R: \mathbb{R}^m \rightarrow \{0,1\}$ be the indicator of this region, and $\tilde{I}_R^\eta: \mathbb{R}^m \rightarrow \mathbb{R}$ is the \textit{FT-mollification} of $I_R$ (you may think of it as a smoothed version of $I_R$ with smoothing parameter $\eta$; see Sections 3 and 4 of \cite{diakonikolas2010bounded} for an overview of FT-mollification, we will only use a few properties and will cite them clearly later). In this language, our desired guarantee is simply a bound on $|\Ex_G[I_R(F(X))] - \Ex_T[I_R(F(X))]|$. Set some parameters $\rho = \eps/m$ and $\eta=m/\rho = m^2/\eps$. The proof of this will entail three steps to show
    \begin{equation*}
        \Ex_G[I_R(F(X))] \approx_{(a)} \Ex_G[\tilde{I}^\eta_R(F(X))] \approx_{(b)} \Ex_T[\tilde{I}^\eta_R(F(X))] \approx_{(c)} \Ex_T[I_R(F(X))].
    \end{equation*}

    These are shown in parts (a), (b), and (c), respectively. 

    \textbf{Part (a). } We will bound $\Ex_G[I_R(F(X))] \approx_{(a)} \Ex_G[\tilde{I}^\eta_R(F(X))]$. This part follows exactly as done in \cite{diakonikolas2010bounded}. We will provide it (almost copied exactly) for completeness. Let $d_2(x,\partial R)$ denote the $L_2$ distance from some point $x \in \mathbb{R}^m$ to the boundary of $R$. Then,

    \begin{align}
        & |\Ex_G[I_R(F(X))] - \Ex_G[\tilde{I}_R^\eta(F(X))]| \label{step:a-begin}\\
        & \le \Ex_G[|I_R(F(X)) - \tilde{I}_R^\eta(F(X))|] \nonumber \intertext{Using Theorem 4.10 of \cite{diakonikolas2010bounded}, which states $|I_R(x) - \tilde{I}_R^\eta(x)| \le \min \left\{1,O\left( \left( \frac{m}{\eta \cdot d_2(x,\partial R)} \right)^2\right) \right\}$:}
        & \le \Pr_G[d_2(F(X),\partial R) \le 2 \rho] + O\left( \sum_{s=1}^\infty \left( \frac{m^2}{\eta^2 2^{2s} \rho^2} \right) \cdot \Pr_G[d_2(F(X),\partial R) \le 2^{s+1} \rho] \right) \nonumber \\
        & \le \sum_{i=1}^m \Pr_G[|F(X)_i - b_i| \le 2\rho] + O\left( \sum_{s=1}^\infty \sum_{i=1}^m 2^{-2s} \cdot \Pr_{G}[|F(X)_i - b_i| \le 2^{s+1}\rho]\right)  \label{step:a-mid} \\
        & \le 2m\rho C_2 + O\left(\sum_{s=1}^{\infty} 2^{-2s} \cdot (2^{s+1}\rho m C_2) \right) \nonumber \tag{by anti-concentration}\\
        & = O(\eps C_2) = O(\eps)\nonumber
    \end{align}

    \textbf{Part (b). } Let us introduce some notation for Taylor expansions around $0$. Let $p_{\le r}$ denote the degree-$r$ Taylor expansion of $\tilde{I}_R^\eta$, let $p_{=r}$ denote the terms of $p_{\le r}$ with degree exactly $r$, and let $p_{>r}(x) \triangleq \tilde{I}_R^\eta(x)-p_{\le r}(x)$. For any polynomial $p$ and multi-index $\alpha$, let $p[\alpha]$ denote the coefficient for this multi-index. We follow the plan of \cite{diakonikolas2010bounded} and decompose into two terms
    \begin{equation*}
        |\Ex_G[\tilde{I}_R^\eta] - \Ex_T[\tilde{I}_R^\eta]| \le |\Ex_G[p_{> k-1}] - \Ex_T[p_{>k-1}]| + |\Ex_G[p_{\le k-1}] - \Ex_T[p_{\le k-1}]|.
    \end{equation*}

    We start by bounding the first term:

    \begin{align*}
        & |\Ex_G[p_{> k-1}] - \Ex_T[p_{>k-1}]|\\
        & \le \Ex_G[|p_{> k-1}|] + \Ex_T[|p_{>k-1}|] \intertext{Using Theorem 4.8 of \cite{diakonikolas2010bounded} to bound $\|\partial^\beta \tilde{I}^\eta_R\|_\infty \le (2 \eta)^{|\beta|}$ for all $\beta \in \mathbb{N}^m$, and Taylor's theorem $|p_{>k-1}(F(x))| \le \sup_{|\beta|=k}\|\partial^\beta \tilde{I}^\eta_R\|_\infty \cdot \frac{\|F(X)\|_1^k}{k!}$:}
        & \le \Ex_G\left[\frac{(2\eta)^k}{k!} \|F(X)\|_1^k\right] + \Ex_T\left[\frac{(2\eta)^k}{k!} \|F(X)\|_1^k\right]\\
        & \le \frac{(2\eta m)^k}{k!} \left( \Ex_G[\max_{i \in [m]}(F(X)_i)^k] + \Ex_T[\max_{i \in [m]}(F(X)_i)^k] \right)\\
        & \le \frac{m(2\eta m)^k}{k!} \left(\max_{i \in [m]} \Ex_G[(F(X)_i)^k] + \max_{i \in [m]}\Ex_T[(F(X)_i)^k] \right)\\
        & \le \frac{m(2\eta m)^k}{k!} \left(2 \sup_{\|v\|_2=1} \Ex_G[\langle X, v \rangle^k] + \sup_{\|v\|_2=1} \left| \Ex_G[\langle X, v \rangle^k] - \Ex_T[\langle X, v \rangle^k] \right| \right)\\
        & \le \frac{m(2\eta m)^k}{k!} \left(2 (C_1 \sqrt{k})^k + \Delta \right)\\
        & \le m \cdot \left(\frac{2e m^3 }{k \eps}\right)^k \cdot \left(2 (C_1 \sqrt{k})^k + \Delta \right)
    \end{align*}
    This quantity is at most $\eps$ if we choose $k \ge C m^6 / \eps^2$ for sufficiently large $C>0$, and $\Delta \le 1$, which are both satisfied by our choice in the theorem statement.

    For the remaining summand, bounding the difference in the expectation of $p_{\le k-1}$ will leverage the lower-order moment bounds as you might expect. The following lemma will be helpful for bounding monomial moments in terms of moments for projections:

    \begin{lemma}[Optimizer of symmetric $k$-linear form. Implied by Equation (2) in \cite{carando2019symmetric}; references proofs in \cite{banach1938homogene,bochnak1971polynomials,pappas2007norm,dineen2012complex}]\label{lemma:symmetric}
        Consider any two random variables $X,Y$ over $\mathbb{R}^d$. Then, 
        \begin{equation*}
            \sup_{\|u_1\|_2,\dots,\|u_k\|_2=1}\Ex_X[\Pi_{j=1}^k \langle X, u_j \rangle ] - \Ex_Y[\Pi_{j=1}^k \langle Y, u_j \rangle ] = \sup_{\|v\|_2=1}  \Ex_X[\langle X,v \rangle^k] - \Ex_Y[\langle Y,v \rangle^k].
        \end{equation*}
    \end{lemma}

    First, we may use this to get a bound in the moment projections for $F(X)$ under $G,H$ with only a blowup depending on $k,m$ (not $d$):

    \begin{claim} \label{claim:f-project}
    For any unit $v \in \mathbb{R}^j$ ($\|v\|_2=1$) and $k' \in \{0,\dots,k\}$,
        \begin{equation*}
            \left| \Ex_G[\langle F(X),v \rangle^k] - \Ex_T[\langle F(X),v \rangle^k] \right| \le m^{k/2} \Delta
        \end{equation*}
    \end{claim}
    \begin{proof} \begin{align*}
            & \left| \Ex_G[\langle F(X),v \rangle^{k'}] - \Ex_T[\langle F(X),v \rangle^{k'}] \right| \\
            & = \left| \Ex_G\left[\left( \sum_{i=1}^m \langle X,a_i \rangle \cdot v_i \right)^{k'}\right] - \Ex_T\left[\left( \sum_{i=1}^m \langle X,a_i \rangle \cdot v_i \right)^{k'}\right] \right|\\
            & = \left| \Ex_G\left[\sum_{i_1,\dots,i_{k'} \in [m]} \Pi_{j=1}^{k'} \langle X, a_{i_j}\rangle \cdot v_{i_j}\right] -   \Ex_T\left[\sum_{i_1,\dots,i_{k'} \in [m]} \Pi_{j=1}^{k'} \langle X, a_{i_j}\rangle \cdot v_{i_j}\right]  \right|\\
            & \le \sum_{i_1,\dots,i_{k'} \in [m]} \left| \Ex_G\left[ \Pi_{j=1}^{k'} \langle X, a_{i_j}\rangle \cdot v_{i_j}\right] - \Ex_T\left[ \Pi_{j=1}^{k'} \langle X, a_{i_j}\rangle \cdot v_{i_j}\right] \right|\\
            & = \sum_{i_1,\dots,i_{k'} \in [m]} \left(\Pi_{j=1}^{k'} |v_{i_j}| \right) \cdot \left| \Ex_G\left[ \Pi_{j=1}^{k'} \langle X, a_{i_j}\rangle\right] - \Ex_T\left[ \Pi_{j=1}^{k'} \langle X, a_{i_j}\rangle \right] \right|\\
            & \le \|v\|_1^{k'}  \sup_{\|u_1\|_2,\dots, \|u_{k'}\|_2=1} \left| \Ex_G[\Pi_{j=1}^{k'} \langle X, u_j \rangle] - \Ex_T[\Pi_{j=1}^{k'} \langle X, u_j \rangle]\right|\\
            & \le m^{{k'}/2} \Delta \, \, \, \qedhere \tag{using \cref{lemma:symmetric}}
        \end{align*}
    \end{proof}
    \begin{corollary}\label{cor:f-moment} For any $k' \in \{0,\dots,k\}$
        \begin{equation*}
            \sup_{i_1,\dots,i_{k'} \in [m]}\left| \Ex_G[\Pi_{j=1}^{k'} f(X)_{i_j}] - \Ex_T[\Pi_{j=1}^{k'} f(X)_{i_j}]   \right| \le m^{{k'}/2} \Delta
        \end{equation*}
    \end{corollary}
    \begin{proof}
        This follows immediately from \cref{lemma:symmetric} and \cref{claim:f-project}.
    \end{proof}

    With these moment bounds in hand, we now bound

    \begin{align}
        & \left| \Ex_G[p_{\le k-1}(F(x))] -  \Ex_T[p_{\le k-1}(F(x))] \right| \\
        & \le k \max_{i \in [k]} \left| \Ex_G[p_{=i}(F(X))] - \Ex_T[p_{=i}(F(X))]\right| \\
        & \le k \max_{i \in [k]} \sum_{\alpha = (j_1,\dots, j_i), j_\ell \in [m]} |p_{=i}[\alpha]| \cdot \left| \Ex_G[F(X)_\alpha] - \Ex_T[F(X)_\alpha] \right| \\
        & \le k \max_{i \in [k]} m^i \cdot \left( \max_{\alpha = (j_1,\dots, j_i), j_\ell \in [m]} |p_{=i}[\alpha]|\right) \cdot \max_{\alpha = (j_1,\dots, j_i), j_\ell \in [m]} \left| \Ex_G[F(X)_\alpha] - \Ex_T[F(X)_\alpha] \right| \intertext{Using Theorem 4.8 of \cite{diakonikolas2010bounded} to bound $\|\partial^\beta \tilde{I}^\eta_R\|_\infty \le (2\eta)^{|\beta|}$ for all $\beta \in \mathbb{N}^m$:}
        & \le k \max_{i \in [k]} m^i \cdot \left( \frac{(2\eta)^i}{i!}\right) \cdot \max_{\alpha = (j_1,\dots, j_i), j_\ell \in [m]} \left| \Ex_G[F(X)_\alpha] - \Ex_T[F(X)_\alpha] \right|\\
        & \le k \max_{i \in [k]} m^i \cdot \left( \frac{(2\eta)^i}{i!}\right) \cdot \left( m^{i/2} \Delta \right)\tag{using \cref{cor:f-moment}}\\
        & \le k \cdot (2m^{3.5}/\eps)^k \Delta \le (4k)^k \Delta \tag{using $k \ge m^6 / \eps^2$}
    \end{align}
    which is at most $\eps$ if we choose $\Delta \le (8k)^{-k}$ as is chosen in the theorem statement.
    
    \textbf{Part (c). } This will follow very similarly to part (a). We may identically argue using the steps from \cref{step:a-begin} to \cref{step:a-mid}. What remains is to show $H$ is sufficiently anti-concentrated in its one-dimensional projections. Note how this is implied if we could show the special case where $H$ fools any pair of $m=2$ halfspaces. One convenient way to prove this, is noting how the result of \cite{gollakota2023moment} was only too lossy for super-constant $m$, so we may use the results after certifying how the conditions hold for moment-matching $F(X)$ under $G$ and $F(X)$ under $H$. We rephrase Theorem 5.6 of \cite{gollakota2023moment} (their result holds for arbitrary functions of the indicators of halfspaces, but for simplicity we just state their result for intersections of halfspaces).

    \begin{theorem}[Theorem 5.6 of \cite{gollakota2023moment}]\label{thm:testable}
        Suppose $D$ and $D'$ are distributions over $\mathbb{R}^d$ such that $D$ satisfies 
        \begin{enumerate}
            \item \textit{$\alpha$-strictly subexponential tails}: For all $\|u \| = 1$, $\Pr_D[|\langle x, u \rangle | > t] \le \exp(-C_1t^{1+\alpha})$ for some positive constant $C_1$. 
            \item \textit{Anticoncentration}: For all $\|u\|=1$ and continuous intervals $T \subset \mathbb{R}$, we have $\Pr_D[\langle x, u \rangle \in T] \le C_2 |T|$ for some positive constant $C_2$.
        \end{enumerate}
        
        Moreover, for some $k \in \mathbb{N}$,  every $j$-moment tuple is close (for $j \in \{0,\dots,k\}$):
        \begin{equation*}
            \left| \Ex_D[\Pi_{\ell=1}^j X_{i_\ell}] - \Ex_D[\Pi_{\ell=1}^j X_{i_\ell}] \right| \le \frac{\sqrt{m}}{2k} \frac{j!}{d^j} \left( \frac{1}{C_5 k^{\alpha/(1+\alpha)}} \right)^{j+1}
        \end{equation*}
        for some positive constant $C_5$.

        Let $f: \mathbb{R}^d \rightarrow \mathbb{R}$ be the indicator of the intersection of $m$ halfspaces. Then, for some constant $C>0$,
        \begin{equation*}
            \left| \Ex_D[f] - \Ex_{D'}[f]\right| \le k^{-\alpha/(1+\alpha)} \sqrt{m} (C \log(\sqrt{m}k^{\alpha/(1+\alpha)}))^{2m}.
        \end{equation*}
    \end{theorem}
    We will invoke this result simply with $d=1$ and $m=2$, using the following immediate corollary:
    \begin{corollary}\label{cor:testable-cor}
        Suppose $D$ and $D'$ are distributions over $\mathbb{R}$ where $D$ has subexponential tails (with $\alpha=1$) and anticoncentration as stated in \cref{thm:testable}. There exists some constant $C^*>0$ where if for some $k \in \mathbb{N}$,  every $j$-moment tuple is close (for $j \in \{0,\dots,k\}$):
        \begin{equation*}
            \left| \Ex_D[\Pi_{\ell=1}^j X_{i_\ell}] - \Ex_{D'}[\Pi_{\ell=1}^j X_{i_\ell}] \right| \le \frac{j!}{k} \left( \frac{1}{C^* \sqrt{k}} \right)^{j+1}
        \end{equation*}
        Then, for any $0<\eps \le \frac{1}{2}$, as long as $k \ge C^* \cdot \frac{1}{\eps^2} \cdot \log^8(1/\eps)$, it holds that
        \begin{equation*}
            \left| \Pr_{D}[x \in [L,R] ] - \Pr_{D'}[x \in [L,R]] \right| \le \eps.
        \end{equation*}
    \end{corollary}
    \begin{proof}
        This is just invoking \cref{thm:testable} with $d=1$ and $m=2$. The moment condition of \cref{thm:testable} follows trivially from the assumption of this corollary. We now simply compute
        \begin{align*}
            &\left| \Pr_{D}[x \in [L,R] ] - \Pr_{D'}[x \in [L,R]] \right| \le k^{-\alpha/(1+\alpha)} \sqrt{m} (C_1 \log(\sqrt{m}k^{\alpha/(1+\alpha)}))^{2m} \\
            &= \frac{\sqrt{2}}{\sqrt{k}} (C_1 \log(\sqrt{2}) + \frac{1}{2}\log(k))^4 \\
            & \le C \cdot  \frac{\log(k)^4}{\sqrt{k}} \\
            & \le C \cdot \left( \log(C^*)^4 + \log(1/\eps)^4 \right) \cdot \left( \frac{\eps}{C^* \log^4(1/\eps)} \right) \tag{by monotonicity and $k \ge \frac{C^* \log^8(1/\eps)}{\eps^2}$} \\
            & \le \eps \quad\qedhere \tag{for sufficiently large $C^*$}
        \end{align*}
    \end{proof}
    We may invoke \cref{cor:testable-cor} on $F(X)_i$ under $G$ and $T$, with $\eps' = \eps/m$. The anticoncentration property holds under $G$ by the assumption of our theorem, and the subexponential tails property holds under $G$ since it is equivalent (up to the choice of constant) to our theorem's subgaussian moment assumption (e.g. see Proposition 2.5.2 of \cite{vershynin2018high}). Using \cref{cor:f-moment}, the moment tuple difference condition holds as long as 
    \begin{align*}
        & m^{k/2} \Delta \le \frac{1}{k} \left(\frac{1}{C^* \sqrt{k}} \right)^{k+1} \\
        & \impliedby \Delta \le \frac{1}{k \cdot m^{k/2}} \left(\frac{1}{C^* \sqrt{k}} \right)^{k+1}\\
        & \impliedby \Delta \le \left( \frac{1}{ C \cdot C^* \cdot k} \right)^{k} \tag{for sufficiently large $C>0$, using $k \ge m,C^* \ge 1$}
    \end{align*}
    which follows from our choice of $\Delta$. Hence, as long as $k \ge C^* \cdot \frac{1}{\eps'^2} \cdot \log^8(1/\eps') = C^* \cdot \frac{m^2}{\eps^2} \cdot \log^8(m/\eps)$, we may conclude, using the steps from \cref{step:a-begin} to \cref{step:a-mid} in part(a):

    \begin{align*}
        & |\Ex_T[I_R(F(X))] - \Ex_T[\tilde{I}_R^\eta(F(X))]| \\
        & \le \sum_{i=1}^m \Pr_T[|F(X)_i - b_i| \le 2\rho] + O\left( \sum_{s=1}^\infty \sum_{i=1}^m 2^{-2s} \cdot \Pr_{H}[|F(X)_i - b_i| \le 2^{s+1}\rho]\right)  \intertext{By \cref{cor:testable-cor}:}
        & \le O(m \eps') + \sum_{i=1}^m \Pr_T[|F(X)_i - b_i| \le 2\rho] + O\left( \sum_{s=1}^\infty \sum_{i=1}^m 2^{-2s} \cdot \Pr_{H}[|F(X)_i - b_i| \le 2^{s+1}\rho]\right)  \\
        & \le O(m \eps') + O(\rho m C_2) = O(\eps C_2) = O(\eps) \quad \qedhere
    \end{align*}
\end{proof}